\documentclass[runningheads]{llncs}

\usepackage[utf8]{inputenc} 
\usepackage[T1]{fontenc}    
\usepackage{hyperref}       
\usepackage{url}            
\usepackage{booktabs}       
\usepackage{amsfonts}       
\usepackage{nicefrac}       
\usepackage{microtype}      
\usepackage[normalem]{ulem} 

\usepackage{graphicx}
\usepackage{tikz}
\usepackage{subcaption}
\usepackage{float}
\usepackage{amsmath}

\usepackage{xcolor}

\usepackage{ifthen}

\newboolean{long}
\setboolean{long}{true}

\newcommand{\longpaper}[2]{\ifthenelse{\boolean{long}}{{#1}}{#2}}

\title{Using Reed-Muller Codes for \\ Classification with Rejection and Recovery}

\author{Daniel Fentham \inst{1} \orcidID{0009-0000-2907-356X} \and David Parker \inst{2} \orcidID{0000-0003-4137-8862} \and Mark Ryan \inst{1} \orcidID{0000-0002-1632-497X}}

\authorrunning{D. Fentham et al.}

\institute{School of Computer Science, University of Birmingham, Birmingham, United Kingdom \email{dxf209@student.bham.ac.uk}, \email{m.d.ryan@cs.bham.ac.uk} \and 
    Department of Computer Science, University of Oxford, Oxford, United Kingdom \email{david.parker@cs.ox.ac.uk}}

\begin{document}
\maketitle
\begin{abstract}

When deploying classifiers in the real world, users expect them to respond to inputs appropriately. However, traditional classifiers are not equipped to handle inputs which lie far from the distribution they were trained on. Malicious actors can exploit this defect by making adversarial perturbations designed to cause the classifier to give an incorrect output. Classification-with-rejection methods attempt to solve this problem by allowing networks to refuse to classify an input in which they have low confidence. This works well for strongly adversarial examples, but also leads to the rejection of weakly perturbed images, which intuitively could be correctly classified. 
To address these issues, we propose Reed-Muller Aggregation Networks (RMAggNet), a classifier inspired by Reed-Muller error-correction codes which can correct and reject inputs. This paper shows that RMAggNet can minimise incorrectness while maintaining good correctness over multiple adversarial attacks at different perturbation budgets by leveraging the ability to correct errors in the classification process. This provides an alternative classification-with-rejection method which can reduce the amount of additional processing in situations where a small number of incorrect classifications are permissible.


\end{abstract}

\section{Introduction}

Deep Neural Networks (DNNs) have shown incredible performance in numerous classification tasks, including image classification \cite{chen2023symbolic}, medical diagnosis \cite{tragakis2023fully} and malware detection \cite{malware}. However, a fundamental shortcoming is that they pass judgement beyond their expertise. When presented with data outside of the distribution they were trained on, DNNs will attempt to classify that data by selecting from one of the finite labels available, often reporting high confidence in the classifications they have made. The most egregious examples of this occur when a DNN is presented with an input which is far outside of the domain it has been trained on (for example, presenting an image of a cat to a text classification model), which it will confidently assign a class to. This behaviour is also present in adversarial examples, which were introduced in the seminal paper by Szegedy et al. in 2018 \cite{szegedy2013intriguing}, where an (almost) invisible perturbation pushes an input far from the training distribution, leading to a confident misclassification \cite{smith2018understanding}. Since then, extensive research has been conducted exploring new, more sophisticated, attacks on networks of different architectures \cite{zou2023universal,morris2020textattack,chen2019shapeshifter} and defences that attempt to mitigate their effectiveness \cite{papernot2016distillation,goodfellow2014explaining,verma2019error}. This results in hesitation when applying DNN models to safety- and security-critical applications where there is a high cost of misclassification.

Classification-with-rejection (CWR) methods \cite{verma2019error,cortes2016learning,charoenphakdee2021classification,song2021error} attempt to address this limitation by refusing to assign a label to an input when the confidence in the classification is low.
In this paper, we present an approach to CWR that parallels ideas from Error-Correcting Output Codes (ECOCs) \cite{verma2019error,song2021error}, where an ensemble of networks perform classification by generating binary strings, extending them with a reject option. ECOC methods have received little attention as a defence mechanism against adversarial attacks, even though the independence of the individual networks offers a natural defence. A notable property of adversarial attacks is that they are highly transferable between models, meaning an adversarial attack crafted for one network will likely deceive another with high probability, provided the networks perform a similar task \cite{papernot2016transferability}. Since ECOC methods promote diversity in the constituent network's tasks, an adversarial attack crafted to change the output reported by one network is less likely to fool another in a way which would result in further misclassification. Moreover, due to the aggregated nature of the resulting classification, an adversary would need to create a perturbation which can fool multiple networks simultaneously, necessitating precise bit-flipping strategies which lead to a valid target class.

In this paper, we introduce Reed-Muller Aggregation Networks (RMAggNet), which apply error correcting codes to the correction and rejection of classifications, ultimately producing a new kind of ECOC classifier. Similar to existing ECOCs, these consist of multiple DNNs, each performing a simple classification task which determines if an input belongs to a defined subset of the classes, resulting in a binary answer. The results of these networks are aggregated together into a binary string which we compare to class binary strings which represent each of the classes from the dataset. If the resulting binary string is the same as a class binary string, we return the associated label as a result, otherwise we attempt to correct the result (if we have a small enough Hamming distance), or reject the result and refuse to classify the input. Thus, unlike existing CWR methods our approach has the ability to both correct and reject inputs. 

We evaluate the effectiveness of RMAggNet by comparing it to two other CWR approaches: an ensemble of networks (with a voting-based rejection process) and Confidence Calibrated Adversarial Training (CCAT)~\cite{Stutz2020ICML}.
After performing tests using the EMNIST and CIFAR-10 datasets with open-box PGD $L_\infty$ and PGD $L_2$ adversarial attacks, we conclude that RMAggNet can greatly reduce the amount of rejected inputs in certain circumstances, making it a viable alternative to methods such as CCAT if some incorrectness is acceptable. \longpaper{}{We expand on this finding using the MNIST dataset, along with closed-box adversarial attacks, in an extended version of this paper \cite{longpaper}.}

In summary, this paper makes the following contributions:

\begin{itemize}
    \item We introduce RMAggNet, a novel ECOC classification method which leverages the power of Reed-Muller codes to create a classifier which can both correct and reject inputs (Section~\ref{sec:method})\footnote{Code available at: \url{https://github.com/dfenth/RMAggNet}}.
    \item \longpaper{We show the effectiveness of RMAggNet on the MNIST, EMNIST and CIFAR-10 datasets with open-box and closed-box gradient and gradient-free adversarial attacks (Sections~\ref{sec:evalmethod} and \ref{sec:results}).}{We show the effectiveness of RMAggNet on the EMNIST and CIFAR-10 datasets with open-box gradient-based adversarial attacks (Sections~\ref{sec:evalmethod} \& \ref{sec:results}).
    }
    \item We discuss the application of RMAggNet to classification tasks, providing guidance on when it may be a strong alternative to other CWR methods (Section~\ref{sec:discussion}).
\end{itemize}

\section{Related work}

\subsection{Error correcting output codes}
Verma and Swami defined an error-correcting output code (ECOC) classification method which uses an ensemble of models, each trained to perform a subset of the classification \cite{verma2019error}. Their model uses Hadamard matrices to construct binary codes, which are assigned to classes from the dataset. Multiple DNNs are then defined to generate a set amount of bits from each code for each class, essentially following a set membership classification approach. When an input is passed to the multiple networks, a vector of real numbers is generated, and the similarity between this vector and the class vectors is calculated, with the most similar vector being returned as the final classification.

The authors argue that this classification method has greater resilience to adversarial attacks than traditional ensemble methods due to the independence of the models, encouraged by the diverse classification tasks. This reduces the chance of multiple coordinated bit-flips occurring due to a single perturbation as a result of the transferability of adversarial attacks.
Verma and Swami focus their attention on multi-bit outputs, where four networks produce a combined total of 16, 32, or 64 bits, encoding the input into 4, 8 or 16 bits, respectively. This approach to ECOC classification leads to similar networks being trained, where, in many cases, the entire set of classes is being used by all networks. This results in each network learning similar features, reducing independence and lowering resilience to transfer attacks.

Song et al., proposed a method which extends the work by Verma and Swami, introducing Error Correcting Neural Networks (ECNN) \cite{song2021error}. This paper improves ECOCs by increasing the number of networks to one per output bit and optimising the codeword matrix using simulated annealing \cite{gamalSA} which encourages each classifier to learn unique features, enhancing robustness against direct and transfer adversarial attacks. However, in practice, the ECNN implementation trains a single network with each classifier having a unique top layer. This reduces the independence between networks since each of them share the same low level features which can be used by adversaries.

These approaches are similar to the method proposed in this paper; however, there are a few key differences. While Verma and Swami \cite{verma2019error} and Song et al. \cite{song2021error} discuss the use of error correction, it is not actively utilised in the classification process. In addition, error correction provides a natural implementation of CWR where outputs which deviate significantly from existing classes can trigger a \textit{reject} option where the classifier refuses to return a result. This paper aims to address these gaps by exploring the application of error correction and classification-with-rejection approaches to ECOC methods. We hope to provide insights into the effectiveness, practicality and benefits of these strategies.

\subsection{Confidence Calibrated Adversarial Training} \label{sec:CCAT}
Many CWR methods have been proposed over the years \cite{cortes2016learning,charoenphakdee2021classification,Stutz2020ICML}. We focus on the Confidence Calibrated Adversarial Training (CCAT) CWR method which was introduced by Stutz et al. in 2020 \cite{Stutz2020ICML}. CCAT attempts to produce a model which is robust to unseen threat models which use different $L_p$ norms or larger perturbations when generating adversarial examples. CCAT achieves good rejection performance through adversarial training where the model is trained to predict the classes of clean data with high confidence and produce a uniform distribution for adversarial examples within an $\epsilon$-ball of the true image. \longpaper{This is extrapolated beyond the $\epsilon$-ball to account for larger perturbations. The label $\tilde{y}$ is generated for adversarial training using equation:
$$
\tilde{y} = \lambda(\delta) \; \text{one-hot}(y)+(1-\lambda(\delta))\frac{1}{|C|}
$$
with adversarial perturbation $\delta$ and function $\lambda$ where $\lambda(\delta) \in [0,1]$ equals 0 for $\|\delta\|_\infty \geq \epsilon$, leading to a uniform distribution, and 1 for $\|\delta\|_\infty = 0$, $C$ is the set of all classes in the dataset, with $|C|$ being the cardinality. This encourages the labels of adversarial examples in the training set to have a uniform distribution, which becomes more uniform as $\delta$ increases. This approach to adversarial training leads to strong generalisation for large $\delta$, which means that perturbed inputs to CCAT are reliably rejected. We produce results using CCAT to act as a comparison with our own method when comparing CWR performance.}{}

\section{Reed-Muller Aggregation Networks (RMAggNet)}\label{sec:method}

\subsection{Reed-Muller codes}
We begin with some brief background on Reed-Muller codes, which are multi-error detecting and correcting codes \cite{mullerRM,reedRM}. This extends earlier work on Hamming codes \cite{hammingCode} and generalise many other error correction methods. 

Reed-Muller is often represented with the notation $[2^m, k, 2^{m-r}]_q$. We set $q$, the number of elements in the finite field, to 2, meaning any codes we create will be binary. In the low-degree polynomial interpretation, $m$ denotes the number of variables and $r$ denotes the highest degree of the polynomial both of which influence the properties of the Reed-Muller code. The first element of the tuple ($2^m$) represents the length of the codewords we will use. The second element ($k$) represents the length of the message we can encode and is calculated as 
\begin{equation}\label{eq:k}
    k=\sum_{i=0}^{r}\binom{m}{i}    
\end{equation}
The final element ($2^{m-r}$) is the minimum Hamming distance between any two codes we generate, and influences the amount of error correction that can be applied. For simplicity, we set $n=2^m$ and $d=2^{m-r}$ condensing the notation to $[n,k,d]_2$.

A key advantage of Reed-Muller codes is that they allow us to unambiguously correct a number of bits equal to the Hamming bound $t$:
\begin{equation}\label{hamming_bound}
    t = \biggl\lfloor \frac{(d-1)}{2} \biggr\rfloor    
\end{equation}
due to the guaranteed Hamming distance between any two codewords. This can be thought of as an open Hamming sphere around each codeword with a radius of $2^{m-r-1}$  which does not intersect any other sphere.

To build Reed-Muller codes with pre-determined Hamming distances, we start by selecting values for $m$ and $r$ which fit our use case, i.e., we can generate codewords of appropriate length with a desired amount of correction. Once we have chosen $m$ and $r$, we can calculate $k$ (see equation \ref{eq:k}) and we can define the low-degree polynomial, which will have $k$ coefficients, $m$ variables and a maximum degree of $r$. This allows us to generate the codewords with a minimal Hamming distance of $d$ between any two of the codes.

We specify the coefficients of the polynomial in $k$ different ways where a single coefficient is set to one, and all others are zero. This gives us $k$ polynomials with fixed coefficients and $m$ free variables. We can then define the basis vectors of the space by instantiating every possible combination of variables for each of the fixed coefficient polynomials. This creates $k$ codewords of length $2^m$ all of which have a guaranteed Hamming distance of at least $d$. We can have up to $2^k$ valid codewords which satisfy the Hamming distance guarantee. These additional codewords can be generated by performing an XOR operation on all possible combinations of the basis vectors generating a closed set. We refer to these binary vectors as \textit{codewords}. 

\longpaper{
For example, if we set $m=3$ and $r=2$, then we can calculate $k = \binom{3}{0} + \binom{3}{1} + \binom{3}{2} = 7$. This means we can generate a polynomial: 
$$
p(z_1, z_2, z_3) = c_0 + c_1z_1 + c_2z_2 + c_3z_3 + c_4z_1z_2 + c_5z_1z_3 + c_6z_2z_3
$$
From this we can generate 7 polynomials with fixed coefficients:
\begin{table}[H]
    \centering
    \begin{tabular}{l}
         $p_{1000000}(z_1, z_2, z_3) = 1$\\ 
         $p_{0100000}(z_1, z_2, z_3) = z_1$\\
         $p_{0010000}(z_1, z_2, z_3) = z_2$\\
         $\vdots$\\
         $p_{0000001}(z_1, z_2, z_3) = z_2z_3$\\
    \end{tabular}
    \label{tab:fixed_coeff_poly}
\end{table}

Since $m=3$ we can have 8 different combinations of variables which we can apply to each polynomial, therefore, we produce the basis vectors:
\begin{table}[H]
    \centering
    \resizebox{0.6\columnwidth}{!}{
    \begin{tabular}{l|c c c c c c c c}
                      & (000) & (001) & (010) & (011) & (100) & (101) & (110) & (111) \\ \hline
        $p_{1000000}$ & 1 & 1 & 1 & 1 & 1 & 1 & 1 & 1\\ 
        $p_{0100000}$ & 0 & 0 & 0 & 0 & 1 & 1 & 1 & 1\\ 
        $p_{0010000}$ & 0 & 0 & 1 & 1 & 0 & 0 & 1 & 1\\ 
        $p_{0001000}$ & 0 & 1 & 0 & 1 & 0 & 1 & 0 & 1\\ 
        $p_{0000100}$ & 0 & 0 & 0 & 0 & 0 & 0 & 1 & 1\\ 
        $p_{0000010}$ & 0 & 0 & 0 & 0 & 0 & 1 & 0 & 1\\ 
        $p_{0000001}$ & 0 & 0 & 0 & 1 & 0 & 0 & 0 & 1\\ 
    \end{tabular}
    }
    \label{tab:basis_vec_table}
\end{table}

Because $d=2^{3-2} = 2$, then each of the basis vectors should have a Hamming distance of at least 2 which is confirmed in the table above.
}{}

\subsection{Reed-Muller Aggregation Networks}\label{sec:rm_aggnet}
We can now define a Reed-Muller Aggregation Network (RMAggNet) which uses multiple networks, trained on separate tasks, to create a binary vector which we can classify, correct, or reject. To create an RMAggNet we start by defining Reed-Muller codes which act as class codewords for the dataset classes we intend to recognise. To define appropriate Reed-Muller codes, we have to consider a number of factors related to the problem we are solving.

The first is the number of classes in the dataset ($|C|$). We must make sure that the message length $k$ is adequate for the number of classes, such that, $|C| \leq 2^k$. From the definition of $k$ (equation \ref{eq:k}) we can see that it depends on both $m$ and $r$, therefore these values are influenced by $|C|$ and must be considered early on in the design process. The number of classes in the dataset is the primary point to consider when deciding on values for $m$ and $r$, because if we do not satisfy $|C| \leq 2^k$ then we will not have an effective classifier.

The second factor we must consider, is whether we have appropriate error correction for the problem. The maximum number of errors we can correct is represented by the Hamming bound $t$ (equation \ref{hamming_bound}) which relies on $d$ which is $2^{m-r}$, so we also need to take this into account when deciding on $m$ and $r$.

The third factor is that we must have a low probability of assigning a valid codeword to a random noise image. A fundamental flaw with traditional DNNs is that they will assign a class to any input, even if the input is far from the distribution they have been trained on. The probability of assigning a random noise image a valid class is $|C|/2^n$ where we have no error correction, however, with error correction, the probability increases to $(|C| \cdot \sum^t_{i=0} \binom{n}{i})/2^n$. This means that it is advantageous to only use the amount of error correction that is necessary for the problem at hand, even if that means we are not correcting the maximum number of bits theoretically possible.

Once we have a set of codewords which fit the problem specification, the length of the codewords ($n$) determines the number of networks included in the aggregation. We can assign each network a unique index value corresponding to an index within the class codeword binary strings. The values at the index positions define a set partition which determines which classes a network is trained to return a 1 for and which it returns a 0 for (i.e., the network returns a 1 for all classes with a 1 at the index and 0 otherwise). We can also view the class codewords as a matrix, with each network assigned to a column with 1s and 0s indicating the partition between sets. By randomly shuffling the class codewords, each network is trained to recognise a set of approximately half of the classes.

With the classification task for each network defined, we can move on to training. The training process requires us to adjust the true labels of each input with each network having a unique set of true labels for the training data, where the true labels correspond to the set partition label. Once the dataset has been adjusted each network is trained independently.

During inference we pass the same input to the $n$ networks which produces $n$ real values $\mathbf{v} \in \mathbb{R}^n$. We select a threshold value $\tau$ which acts as a bias, with a large $\tau$ leading to codewords consisting of more 0 bits. We compare each of the $n$ real values of $\mathbf{v}$ to $\tau$ with the following rule:
$$
v_i = 
    \begin{cases} 
        1 & \text{if } v_i \geq \tau\\
        0 & \text{otherwise}
    \end{cases}
$$

This produces a binary string which we can compare to the class codewords. If any of the class codewords match the predicted binary string exactly, we can return the label associated with it as the result; however, if none match, we calculate the Hamming distance between the prediction and the class codewords. If we find a Hamming distance less than or equal to $t$, then we can unambiguously correct to, and return, that class codeword due to the properties of Reed-Muller codes. Otherwise, we refuse to classify the input and return a rejection.

\section{Evaluation methodology}\label{sec:evalmethod}

\subsection{Threat model}
We begin by establishing the threat model under which we expect the RMAggNet defence to operate effectively as per the recommendations made by Carlini et al. \cite{carlini2019evaluating}. Our assumptions consider an adversary who knows the purpose of the model (i.e., the classes the model can output) and is capable of providing the model with inputs. The actions of the adversary are constrained by a limited perturbation cost, where the $L_p$-norm between the original ($x$) and perturbed ($\tilde{x}$) image must be below some threshold $\epsilon$, i.e., ($|x-\tilde{x}|_{p} \leq \epsilon$), where $p$ is either $2$ or $\infty$ depending on the attack used. The norm used for an attack changes the focus of the adversarial perturbation. The $L_2$ attacks encourage small changes across all input dimensions, which distributes the perturbation across multiple features, whereas the $L_\infty$ attack encourages perturbations which focus on a single feature, maximising this as much as the budget ($\epsilon$) allows. The ultimate aim of the adversary is to generate perturbed images which are not noticeable to a time constrained human.

The level of access to the model granted to the adversary depends on the specific adversarial attack being employed. In the case of open-box attacks, the adversary has full access to the target model and can generate adversarial perturbations tailored to deceive that particular network. This represents the worst-case scenario. On the other hand, closed-box attacks represent a more realistic setting where the adversary does not have access to the model parameters. In this case it is necessary to train a surrogate model which performs the same classification task as the target model. Adversarial examples are then generated for this surrogate model, which leverage transferability to create adversarial examples for the target model. \longpaper{Both of these attacks will be used in the following experiments.}{In this paper we focus on the more challenging open-box attacks. Experiments using closed-box attacks can be found in the extended version of this paper \cite{longpaper}.}

\longpaper{We employ a range of attacks to generate adversarial examples, including gradient-based attacks such as Projected Gradient Descent (PGD) \cite{madry2017towards}, both in the $L_2$- and $L_\infty$-norm, and the gradient free Boundary attack in the $L_2$-norm \cite{brendel2018decisionbased}.We also use these adversarial generation methods for the transfer attacks to demonstrate the robustness of these approaches in a closed-box setting. The use of both gradient and gradient-free attacks allows us to more thoroughly evaluate the robustness of the models. While gradient based attacks tend to produce stronger adversarial examples, they can fail to produce effective perturbations if the target model performs any kind of gradient masking. To ensure that we have a fair and reliable evaluation of the robustness we include gradient-free attacks to eliminate the possibility that any results are solely due to masking the model gradients.}
{We employ two attacks to generate adversarial examples, focusing on the gradient-based Projected Gradient Descent (PGD) attack \cite{madry2017towards}, using both the $L_2$- and $L_\infty$-norm. In the extended paper we also include the gradient-free Boundary attack in the $L_2$-norm \cite{brendel2018decisionbased}, and transfer attacks to demonstrate the adversarial robustness of these approaches in a closed-box setting. The use of both gradient and gradient-free attacks allows us to more thoroughly evaluate the robustness of the models. While gradient based attacks tend to produce stronger adversarial examples, they can fail to produce effective perturbations if the target model performs any kind of gradient masking. To ensure that we have a fair and reliable evaluation of the robustness we include gradient-free attacks to eliminate the possibility that any results are solely due to masking the model gradients.}

\subsection{Comparison methods}\label{sec:comparison_methods}
To evaluate the classification and rejection ability of RMAggNet we have implemented two other comparison methods. 

The first is a traditional ensemble method which consists of $n$ networks (where $n$ is the same number of networks used for RMAggNet) each of which are trained to perform the full classification task, as opposed to RMAggNet where each network is trained to perform set membership over two partitions. To aggregate the ensemble method results from the multiple networks, we have set up a simple voting system with an associated threshold ($\sigma$). When an input is passed to the ensemble, each network classifies the data producing a predicted class. If we exceed the threshold with the percentage of networks that agree on a single class, that class is returned as the most likely answer, otherwise, the input is rejected and no class is returned. 

The second is the CCAT method (see Section~\ref{sec:CCAT}) which uses adversarial training as per the original paper \cite{Stutz2020ICML} using the original code which is slightly modified \footnote{Available: \url{https://github.com/davidstutz/confidence-calibrated-adversarial-training}}. The adversarial training process allows CCAT to reject adversarial inputs within an $\epsilon$-ball by learning to return a uniform distribution over all of the classes. This is then extrapolated beyond the $\epsilon$-ball to larger perturbations. A threshold ($\tau$) is specified which represents the confidence bound that must be exceeded so that the result is not rejected. Unlike the original paper where an optimal $\tau$ is calculated based on performance on the clean dataset, we vary $\tau$ to determine the effect on the rejection ability. 

\subsection{Datasets}
We use multiple datasets to evaluate the effectiveness of RMAggNet on a variety of classification tasks. We focus on the \longpaper{MNIST, }{}EMNIST (balanced) \cite{cohen2017emnist} and CIFAR-10 datasets. \longpaper{The MNIST dataset provides us with the simplest classification task with grey-scale images of size $28 \times 28$ of 10 possible classes. This is extended by EMNIST}{The EMNIST dataset provides us with a simple classification task} which consists of 131,600 grey-scale images of size $28 \times 28$, with 47 balanced classes including handwritten digits, upper- and lower-case letters (with some lower case classes excluded). Since we have 47 classes, the number of networks in RMAggNet is expanded to 32, with the same amount used for the Ensemble method. CIFAR-10 represents a more challenging classification task, increasing the image complexity with full colour images of size $32 \times 32$ over 10 possible classes which uses 16 networks for RMAggNet and Ensemble. \longpaper{We also use two extra datasets for out-of-distribution tests which are Fashion MNIST (FMNIST) \cite{xiao2017fmnist}, which consists of $28 \times 28$ images of articles of clothing from 10 classes, and a dataset of uniform random noise images.}

\subsection{Generation of adversarial examples} \label{sec:adv_gen}
To generate adversarial images from the selected datasets we use the FoolBox library \cite{rauber2017foolboxnative,rauber2017foolbox}. We generate adversarial images using \longpaper{PGD $L_2$, PGD $L_\infty$ and Boundary $L_2$ attacks}{PGD $L_2$ and PGD $L_\infty$ attacks}. Due to the complex nature of some of the networks, adjustments needed to be made to generate adversarial examples.

\textbf{RMAggNet}: Due to the non-differentiable nature of RMAggNet from thresholding, direct attacks are difficult to generate. Following approaches such as BPDA \cite{athalye2018obfuscated} we implement a hybrid RMAggNet which replaces the final mapping from a binary string to class (or reject) with a Neural Network. This allows us to backpropagate through the entire model to produce effective adversarial examples.

\textbf{Ensemble}: We implement an ensemble via logits method \cite{dong2018boosting} where the result of each network is weighted. Due to the voting system for the rejection we set equal weights over all networks. This approach allows us to have an ensemble method which mimics the voting output, except it is differentiable, therefore we can generate adversarial examples using the multiple networks directly.

\textbf{CCAT}: Since CCAT is a standard Network which has undergone specific adversarial training, the generation of adversarial attacks is simple. Many attacks in the FoolBox library are able to generate adversarial examples without any modification of the network.

\section{Results}\label{sec:results}

\longpaper{
\subsection{Reed-Muller hyperparameters} \label{sec:results_hyperparams}
Our first experiment explores the use of Reed-Muller hyperparameters. Reed-Muller codes provides us with fine-grained control when it comes to the codeword space. Increasing the number of variables ($m$) gives us access to a larger space at the cost of an increased number of networks ($2^m$), whereas increasing the degree of the polynomials ($r$) allows us to use more of this space at the cost of error correction and an increase in the probability an out-of-distribution input will be randomly assigned a valid class. To explore the effects these parameters have on the classification ability of RMAggNet, both for in- and out-of-distribution data we set up a number of experiments which vary the $m$ and $r$, effectively altering the number of networks used for the aggregated classification, and the amount of error correction we allow.

The results of these experiments can be seen in tables \ref{tab:RMA_m4r1}, \ref{tab:RMA_m4r2}, \ref{tab:RMA_m5r1}. This data is generated by testing three different versions of RMAggNet: ($m=4$, $r=1$) expressed as $[16,5,8]_2$, ($m=4$, $r=2$) expressed as $[16,11,4]_2$, and ($m=5$, $r=1$) expressed as $[32,6,16]_2$, with potential error correction of 3, 1 and 7 bits respectively. The tests are performed over three datasets, a clean MNIST dataset, a dataset consisting of random uniform noise (Noise), and Fashion MNIST (FMNIST) \cite{xiao2017fmnist}. The last two act as out-of-distribution datasets where the Noise dataset has no semantic meaning, and FMNIST has semantic meaning, but should still be considered far from the distribution the models were trained on (and therefore, should be rejected). For all experiments going forward we set the threshold to $\tau = 0.5$ when generating codewords so that there is no bias towards a particular bit.

On the clean dataset $[16,5,8]_2$ and $[16,11,4]_2$ have very similar performance at $EC=1$, but $[16,5,8]_2$, with the larger amount of error correction, can achieve better accuracy. Increasing accuracy through error correction intuitively makes sense since we are effectively expanding the Hamming sphere around each class codeword, including more codewords which can be considered valid. However, by expanding the Hamming sphere, we also increase the space of erroneously assigned codewords which can be assigned a class incorrectly. From $[32,6,16]_2$ we can clearly see that earlier $EC$ values transform far more rejections to correct classifications, however, this trend diminishes as we allow more correction. For example, if we focus on different $EC$ values in $[32,6,16]_2$, increasing the amount of $EC$ from 0 to 1, the performance on MNIST (Clean) changes from $(85.29, 14.68, 0.03)$ to $(96.58, 3.38, 0.04)$, which means that by correcting just one bit, we are able to see changes of $(+11.29, -12.40, +0.01)$ where a majority of the rejections are transferred to correct classifications. However, as we continue increasing the $EC$, the conversion rate from rejections to correct classifications decreases. As we move from $EC=6$ to $EC=7$ the difference in correctness, rejections and incorrectness becomes $(+0.21, -0.33, +0.12)$ where approximately one-third of the rejected inputs are transferred to an incorrect classification.

Overall $[32,6,16]_2$ achieves better accuracy at its maximum $EC$ than $[16,5,8]_2$ with only a slight increase in incorrect classifications. This is likely due to the aggregation of information from more networks, which is a common occurrence in ensemble based methods. However, on the clean dataset at $EC=0$, the difference in the number of incorrect classifications is negligible, while the number of correct classifications for $[16,5,8]_2$ is much higher. It is worth noting that we may observe the law of diminishing returns here as a doubling in the number of networks used for aggregation (16 for $[16,5,8]_2$ and 32 for $[32,6,16]_2$) leads to a 0.21\% increase in accuracy with a 0.1\% increase in the number of incorrect classifications. This strongly indicates that the $[16,5,8]_2$ model has an ideal balance between classification ability and computational cost.

On the out-of-distribution datasets (Noise and FMNIST), all models reliably reject the noise data better than FMNIST which is to be expected since FMNIST has a semantic structure which could share features with data the model is trained on. $[32,6,16]_2$ has excellent performance on both out-of-distribution datasets at $EC=0$, far above $[16,5,8]_2$ and $[16,11,4]_2$. This is likely due to the probability of a random input being assigned a class codeword which is much lower for $[32,6,16]_2$ at $2.33 \times 10^{-9}$ compared to $1.53 \times 10^{-4}$ for both $[16,5,8]_2$ and $[16,11,4]_2$. 

\begin{table}[htbp]
    \caption{Results for RMAggNet over multiple values of $m$, $r$. Three datasets are used, MNIST, Noise and FMNIST,  where Noise and FMNIST show the performance on out-of-distribution data which is semantically random and structured respectively. EC denotes the amount of error correction. For MNIST (Clean) higher correctness is better, for Noise and FMNIST, higher rejection is better.}
    
    \begin{subtable}{\textwidth}
        \centering
        \resizebox{1\columnwidth}{!}{
        \begin{tabular}{|*{9}{c|}}
            \hline
            \multicolumn{9}{|l|}{\textbf{RMAggNet} $[16,5,8]_2$}\\
            \hline
            & & \multicolumn{3}{|c|}{\textbf{MNIST (Clean)}} & \multicolumn{2}{|c|}{\textbf{Noise}} & \multicolumn{2}{|c|}{\textbf{FMNIST}}\\
            \hline
            \textbf{EC} & $\tau$ & \textbf{Correct} & \textbf{Rejected} & \textbf{Incorrect} & \textbf{Rejected} & \textbf{Incorrect} & \textbf{Rejected} & \textbf{Incorrect}\\
            \hline
            0 & 0.5 & \textbf{96.08} & 3.85 & 0.07 & \textbf{99.39} & 0.61  & \textbf{94.01} & 5.99    \\
            1 & 0.5 & \textbf{97.80} & 2.11 & 0.09 & \textbf{94.97} & 5.03  & \textbf{84.00} & 16.00   \\
            2 & 0.5 & \textbf{98.75} & 1.01 & 0.24 & \textbf{82.10} & 17.90 & \textbf{65.35} & 34.65  \\
            3 & 0.5 & \textbf{99.14} & 0.46 & 0.40 & \textbf{51.07} & 48.93 & \textbf{34.00} & 66.00 \\
            \hline
     
        \end{tabular}
        }
        \caption{Results for $m=4$, $r=1$.}
        \label{tab:RMA_m4r1}
    \end{subtable}

    \begin{subtable}{\textwidth}
        \centering
        \resizebox{1\columnwidth}{!}{%
        \begin{tabular}{|*{9}{c|}}
            \hline
            \multicolumn{9}{|l|}{\textbf{RMAggNet} $[16,11,4]_2$}\\
            \hline
            & & \multicolumn{3}{|c|}{\textbf{MNIST (Clean)}} & \multicolumn{2}{|c|}{\textbf{Noise}} & \multicolumn{2}{|c|}{\textbf{FMNIST}}\\
            \hline
            \textbf{EC} & $\tau$ & \textbf{Correct} & \textbf{Rejected} & \textbf{Incorrect} & \textbf{Rejected} & \textbf{Incorrect} & \textbf{Rejected} & \textbf{Incorrect}\\
            \hline
            0 & 0.5 & \textbf{95.89} & 4.06 & 0.05 & \textbf{99.30} & 0.70 & \textbf{96.23} & 3.77 \\
            1 & 0.5 & \textbf{97.92} & 1.93 & 0.15 & \textbf{95.02} & 4.98 & \textbf{89.48} & 10.52 \\
            \hline
            
        \end{tabular}
        }
        \caption{Results for $m=4$, $r=2$.}
        \label{tab:RMA_m4r2}
    \end{subtable}

    \begin{subtable}{\textwidth}
        \centering
        \resizebox{1\columnwidth}{!}{
        \begin{tabular}{|*{9}{c|}}
            \hline
            \multicolumn{9}{|l|}{\textbf{RMAggNet} $[32,6,16]_2$}\\
            \hline
            & & \multicolumn{3}{|c|}{\textbf{MNIST (Clean)}} & \multicolumn{2}{|c|}{\textbf{Noise}} & \multicolumn{2}{|c|}{\textbf{FMNIST}}\\
            \hline
            \textbf{EC} & $\tau$ & \textbf{Correct} & \textbf{Rejected} & \textbf{Incorrect} & \textbf{Rejected} & \textbf{Incorrect} & \textbf{Rejected} & \textbf{Incorrect}\\
            \hline
            0 & 0.5 & \textbf{85.29} & 14.68 & 0.03 & \textbf{100.00} & 0.00  & \textbf{99.03} & 0.97   \\
            1 & 0.5 & \textbf{96.58} & 3.38  & 0.04 & \textbf{100.00} & 0.00  & \textbf{96.38} & 3.62   \\
            2 & 0.5 & \textbf{97.70} & 2.22  & 0.08 & \textbf{99.97}  & 0.03  & \textbf{92.04} & 7.96   \\
            3 & 0.5 & \textbf{98.31} & 1.57  & 0.12 & \textbf{99.81}  & 0.19  & \textbf{85.16} & 14.84  \\
            4 & 0.5 & \textbf{98.66} & 1.15  & 0.19 & \textbf{99.12}  & 0.88  & \textbf{76.00} & 24.00  \\
            5 & 0.5 & \textbf{98.88} & 0.82  & 0.30 & \textbf{96.57}  & 3.43  & \textbf{65.42} & 34.58  \\
            6 & 0.5 & \textbf{99.14} & 0.48  & 0.38 & \textbf{89.94}  & 10.06 & \textbf{53.36} & 46.64  \\
            7 & 0.5 & \textbf{99.35} & 0.15  & 0.50 & \textbf{76.71}  & 23.29 & \textbf{39.30} & 60.70  \\
            \hline
            
        \end{tabular}%
        }
        \caption{Results for $m=5$, $r=1$.}
        \label{tab:RMA_m5r1}
    \end{subtable}
    
\end{table}
}{}

\longpaper{
\subsection{MNIST Dataset} \label{sec:res_mnist}
Using the hyperparameters found in the previous section (\ref{sec:results_hyperparams}) we construct a RMAggNet model for the MNIST dataset with $m=4$, $r=1$ giving us 16 networks with 3 bits of error correction ($[16,5,8]_2$). We define an equal number of ensemble networks for parity. While the $[32,6,16]_2$ network is able to convincingly outperform it on out-of-distribution data, it has negligible improvement in correctness on the clean dataset and requires twice the number of networks compared to $[16,5,8]_2$, therefore, for computational cost we use 16 networks. The network architecture is shown in table \ref{tab:mnist_arch}.

\begin{table}[htbp]
    \caption{Model architecture for classifying MNIST data. The $c$ parameter is for the varying output sizes depending on the system i.e. $c=1$ for RMAggNet, $c=10$ for Ensemble and CCAT.}
    \label{tab:mnist_arch}
    \centering
    \begin{tabular}{r|l}
        \textbf{Layer} & \textbf{Parameters} \\
        \hline
        Conv2D & channels = 32, kernel size = $5 \times 5$, padding = 2 \\
        ReLU &  \\
        MaxPool2D & pool size = $2\times 2$ \\
        Conv2D & channels = 64, kernel size = $5 \times 5$, padding = 2 \\
        ReLU &  \\
        MaxPool2D & pool size = $2\times 2$ \\
        Flatten & \\
        Linear & in = $7 \times 7 \times 64$, out = 1024 \\
        ReLU & \\
        Linear & in = 1024, out = $c$ \\
    \end{tabular}
    
\end{table}

Table \ref{tab:mnist_res_clean} reports the results on the clean MNIST dataset where we aim to maximise correctness. All three methods report similar correctness with CCAT performing slightly better. If we compare equivalent correctness over all methods ($EC=3$ for RMAggNet, $\sigma=0.4$ for Ensemble and $\tau=0.4$ for CCAT) we can see that RMAggNet is able to report the lowest incorrectness value. This indicates that RMAggNet is able to use the rejection option more effectively to reduce the number of incorrect classifications being made.

Tables \ref{tab:noise_mnist} and \ref{tab:ood_mnist} show the results of all architectures applied to out-of-distribution datasets, with the aim of maximising rejection. Both sets of results show that all architectures are able to reject a large majority of the out-of-distribution data at their most conservative settings (Ensemble $\tau=1.0$ and RMAggNet $EC=0$). CCAT is the highest performing model, rejecting nearly 100\% of the data at nearly all thresholds, excluding $\tau=0$ and $\tau=0.1$ where the CCAT confidence threshold is effectively ignored. RMAggNet shows slightly better performance over Ensemble in table \ref{tab:noise_mnist}, however, Ensemble outperforms RMAggNet in the more challenging Fashion MNIST dataset (table \ref{tab:ood_mnist}) which, unlike the uniform noise dataset, contains low-level features which can be shared with the MNIST dataset. This could impact RMAggNets performance more than Ensemble because Ensemble performs full classification, therefore later layers in the network will contain high-level features which are able to distinguish between the classes it is trained on at a higher level of abstraction. However, RMAggNet only has to perform set membership checks on the inputs, therefore lower-level features may be learned by the networks, meaning that the lower-level features present in FMNIST may have a greater influence over RMAggNet as it attempts to generalise.

\begin{table}[htbp]
    \caption{Percentages of correct, rejected, and incorrect classifications for the clean MNIST dataset. Both Ensemble and RMAggNet models consist of 16 networks. Higher correctness is better.} 
    
    \label{tab:mnist_res_clean}
\begin{subtable}{0.32938\columnwidth}
        \resizebox{1\columnwidth}{!}{
        \begin{tabular}{|*{4}{c|}}
            \hline
            \multicolumn{4}{|l|}{\textbf{CCAT}}\\
            \hline
            \textbf{$\tau$} & \textbf{Correct} & \textbf{Rejected} & \textbf{Incorrect} \\
            \hline
            0    & \textbf{99.49} & 0.00 & 0.51 \\
            0.10 & \textbf{99.49} & 0.00 & 0.51 \\
            0.20 & \textbf{99.46} & 0.04 & 0.50 \\
            0.30 & \textbf{99.43} & 0.10 & 0.47 \\
            0.40 & \textbf{99.38} & 0.17 & 0.45 \\
            0.50 & \textbf{99.28} & 0.32 & 0.40 \\
            0.60 & \textbf{99.16} & 0.50 & 0.34 \\
            0.70 & \textbf{99.01} & 0.68 & 0.31 \\
            0.80 & \textbf{98.87} & 0.89 & 0.24 \\
            0.90 & \textbf{98.41} & 1.41 & 0.18 \\
            1.0  & \textbf{47.41} & 52.59 & 0.00 \\
            \hline
        \end{tabular}
        }
        \label{tab:mnist_ccat_clean}
    \end{subtable}
    \begin{subtable}{0.32938\columnwidth}
        \resizebox{1\columnwidth}{!}{
        \begin{tabular}{|*{4}{c|}}
            \hline
            \multicolumn{4}{|l|}{\textbf{Ensemble}}\\
            \hline
            \textbf{$\sigma$} & \textbf{Correct} & \textbf{Rejected} & \textbf{Incorrect} \\
            \hline
            0    & \textbf{99.34} & 0.00 & 0.66 \\
            0.10 & \textbf{99.34} & 0.00 & 0.66 \\
            0.20 & \textbf{99.34} & 0.00 & 0.66 \\
            0.30 & \textbf{99.34} & 0.00 & 0.66 \\
            0.40 & \textbf{99.34} & 0.00 & 0.66 \\
            0.50 & \textbf{99.33} & 0.04 & 0.63 \\
            0.60 & \textbf{99.23} & 0.19 & 0.58 \\
            0.70 & \textbf{98.98} & 0.60 & 0.42 \\
            0.80 & \textbf{98.81} & 0.84 & 0.35 \\
            0.90 & \textbf{98.50} & 1.22 & 0.28 \\
            1.0  & \textbf{97.27} & 2.62 & 0.11 \\
            \hline
        \end{tabular}
        }
        \label{tab:mnist_ensemble_clean}
    \end{subtable}    
    \begin{subtable}{0.32938\columnwidth}
        \resizebox{1\columnwidth}{!}{
        \begin{tabular}{|*{4}{c|}}
            \hline
            \multicolumn{4}{|l|}{\textbf{RMAggNet} $[16,5,8]_2$}\\
            \hline
            \textbf{EC} & \textbf{Correct} & \textbf{Rejected} & \textbf{Incorrect} \\
            \hline
            0 & \textbf{96.70} & 3.24 & 0.06 \\
            1 & \textbf{98.38} & 1.50 & 0.12 \\
            2 & \textbf{99.02} & 0.78 & 0.20 \\
            3 & \textbf{99.34} & 0.27 & 0.39 \\
            \hline
        \end{tabular}
        }
        \label{tab:mnist_rmaggnet_clean}
    \end{subtable}
\end{table}

\begin{table}[htbp]
    \caption{Percentages of rejected and incorrect classifications on a dataset consisting of uniform random noise images for all models trained on the MNIST dataset. Lower incorrectness is better.}
    \label{tab:noise_mnist}
    \begin{subtable}[h]{0.32938\textwidth}
        \resizebox{1\columnwidth}{!}{
        \begin{tabular}{|*{3}{c|}}
            \hline
            \multicolumn{3}{|l|}{\textbf{CCAT}}\\
            \hline
            \textbf{$\tau$} & \textbf{Rejected} & \textbf{Incorrect} \\
            \hline
            0    & 0.00   & \textbf{100.00}  \\
            0.10 & 0.00   & \textbf{100.00}  \\
            0.20 & 100.00 & \textbf{0.00}    \\
            0.30 & 100.00 & \textbf{0.00}    \\
            0.40 & 100.00 & \textbf{0.00}    \\
            0.50 & 100.00 & \textbf{0.00}    \\
            0.60 & 100.00 & \textbf{0.00}    \\
            0.70 & 100.00 & \textbf{0.00}    \\
            0.80 & 100.00 & \textbf{0.00}    \\
            0.90 & 100.00 & \textbf{0.00}    \\
            1.0  & 100.00 & \textbf{0.00}    \\
            \hline
        \end{tabular}
        }
        \label{tab:noise_mnist_ccat}
    \end{subtable}
    \begin{subtable}[h]{0.32938\textwidth}
    \resizebox{1\columnwidth}{!}{
        \begin{tabular}{|*{3}{c|}}
            \hline
            \multicolumn{3}{|l|}{\textbf{Ensemble}}\\
            \hline
            \textbf{$\sigma$} & \textbf{Rejected} & \textbf{Incorrect} \\
            \hline
            0    & 0.00  & \textbf{100.00} \\
            0.10 & 0.00  & \textbf{100.00} \\
            0.20 & 0.00  & \textbf{100.00} \\
            0.30 & 0.01  & \textbf{99.99}  \\
            0.40 & 8.25  & \textbf{91.75}  \\
            0.50 & 41.61 & \textbf{58.39}  \\
            0.60 & 60.34 & \textbf{39.66}  \\
            0.70 & 84.32 & \textbf{15.68}  \\
            0.80 & 90.71 & \textbf{9.29}   \\
            0.90 & 97.64 & \textbf{2.36}   \\
            1.0  & 99.78 & \textbf{0.22}   \\
            \hline
        \end{tabular}
        }
        \label{tab:noise_mnist_ensemble}
    \end{subtable}
    \begin{subtable}[h]{0.32938\textwidth}
    \resizebox{1\columnwidth}{!}{
        \begin{tabular}{|*{3}{c|}}
            \hline
            \multicolumn{3}{|l|}{\textbf{RMAggNet} $[16,5,8]_2$}\\
            \hline
            \textbf{EC} & \textbf{Rejected} & \textbf{Incorrect} \\
            \hline
            0 & 99.88 & \textbf{0.12} \\
            1 & 98.58 & \textbf{1.42} \\
            2 & 91.81 & \textbf{8.19} \\
            3 & 67.14 & \textbf{32.86}\\
            \hline
        \end{tabular}
        }
        \label{tab:noise_mnist_rmaggnet}
    \end{subtable}
    
\end{table}
\begin{table}[htbp]
    \caption{Percentages of rejected and incorrect classifications on the Fashion MNIST dataset for all models trained on the MNIST dataset. Lower incorrectness is better.}
    \label{tab:ood_mnist}
    \begin{subtable}[h]{0.32938\textwidth}
        \resizebox{1\columnwidth}{!}{
        \begin{tabular}{|*{3}{c|}}
            \hline
            \multicolumn{3}{|l|}{\textbf{CCAT}}\\
            \hline
            \textbf{$\tau$} & \textbf{Rejected} & \textbf{Incorrect} \\
            \hline
            0    & 0.00   & \textbf{100.00} \\
            0.10 & 0.00   & \textbf{100.00} \\
            0.20 & 99.02  & \textbf{0.98}   \\
            0.30 & 99.50  & \textbf{0.50}   \\
            0.40 & 99.68  & \textbf{0.32}   \\
            0.50 & 99.77  & \textbf{0.23}   \\
            0.60 & 99.85  & \textbf{0.15}   \\
            0.70 & 99.90  & \textbf{0.10}   \\
            0.80 & 99.91  & \textbf{0.09}   \\
            0.90 & 99.94  & \textbf{0.06}   \\
            1.0  & 100.00 & \textbf{0.00}   \\
            \hline
        \end{tabular}
        }
        \label{tab:ood_mnist_ccat}
    \end{subtable}
    \begin{subtable}[h]{0.32938\textwidth}
        \resizebox{1\columnwidth}{!}{
        \begin{tabular}{|*{3}{c|}}
            \hline
            \multicolumn{3}{|l|}{\textbf{Ensemble}}\\
            \hline
            \textbf{$\sigma$} & \textbf{Rejected} & \textbf{Incorrect} \\
            \hline
            0    & 0.00   & \textbf{100.00} \\
            0.10 & 0.00   & \textbf{100.00} \\
            0.20 & 0.00   & \textbf{100.00} \\
            0.30 & 3.03   & \textbf{96.97}  \\
            0.40 & 62.13  & \textbf{37.87}  \\
            0.50 & 94.52  & \textbf{5.48}   \\
            0.60 & 98.73  & \textbf{1.27}   \\
            0.70 & 99.95  & \textbf{0.05}   \\
            0.80 & 99.99  & \textbf{0.01}   \\
            0.90 & 100.00 & \textbf{0.00}   \\
            1.0  & 100.00 & \textbf{0.00}   \\
            \hline
        \end{tabular}
        }
        \label{tab:ood_mnist_ensemble}
    \end{subtable}    
    \begin{subtable}[h]{0.32938\textwidth}
        \resizebox{1\columnwidth}{!}{
        \begin{tabular}{|*{3}{c|}}
            \hline
            \multicolumn{3}{|l|}{\textbf{RMAggNet} $[16,5,8]_2$}\\
            \hline
            \textbf{EC} & \textbf{Rejected} & \textbf{Incorrect} \\
            \hline
            0 & 96.07 & \textbf{3.93}  \\
            1 & 88.21 & \textbf{11.79} \\
            2 & 74.47 & \textbf{25.53} \\
            3 & 48.13 & \textbf{51.87} \\
            \hline
        \end{tabular}
        }
        \label{tab:ood_mnist_rmaggnet}
    \end{subtable}
\end{table}

Table \ref{tab:mnist_transfer_pgdlinf} shows the performance of the PGD $L_\infty$ transfer attack on the CCAT, Ensemble, and RMAggNet methods, generated from a surrogate model which shares the same architecture as a single model from the Ensemble method. We use the performance of the surrogate model (table \ref{tab:mnist_surrogate_pgdlinf_res}) as a guide to determine the appropriate perturbation budget. In this setting we aim to minimise incorrectness. The CCAT results show that at $\tau > 0.1$ we achieve 0\% incorrectness with all adversarial inputs being rejected at all $\epsilon$, even though at lower $\epsilon$, 91\% of inputs could still be correctly classified. RMAggNet has higher incorrectness than CCAT, but shows strong classification ability for $\epsilon \in \{0.05, 0.10\}$ with a similar correctness to Ensemble at $\epsilon=0.05$, with more than half the number of incorrect classifications. This becomes even more pronounced at $\epsilon=0.10$ where RMAggNet performs over 10\% better with approximately one-third the number of incorrect classifications, which can be reduced even further while still achieving a higher correctness. However, at $\epsilon = 0.30$ both Ensemble and RMAggNet have high incorrectness compared to CCAT, increasing to a minimum of 92.50\% and 43.50\% respectively.

\begin{table}[htbp]
    \caption{Results for the transfer attacks using PGD $L_\infty$. Table \ref{tab:mnist_surrogate_pgdlinf_res} shows the accuracy of the surrogate model on the PGD $L_\infty$ adversarial datasets. Table \ref{tab:mnist_transfer_pgdlinf} shows the results of the adversarial datasets on the CCAT, Ensemble and RMAggNet models.} 
    \centering
    \begin{subtable}[htbp]{\textwidth}
        \caption{Accuracy of the surrogate MNIST classifier on the adversarial datasets generated using PGD $L_\infty$ with different perturbation budgets ($\epsilon$).}
        \label{tab:mnist_surrogate_pgdlinf_res}
        \begin{center}
            \begin{tabular}{|c|c|}
                \hline
                 $\epsilon$ & \textbf{Accuracy (\%)} \\
                 \hline
                 0.00 & 99.10 \\
                 0.05 & 93.90 \\
                 0.1  & 63.90 \\
                 0.3  & 0.00 \\
                 \hline
            \end{tabular}
        \end{center}
    \end{subtable}
    \begin{subtable}[htbp]{\textwidth}
    \caption{Percentage of correct, rejected and incorrect classifications of the models using transfer attacks on a surrogate MNIST classifier using the PGD $L_\infty$ attack. Lower incorrectness is better.}
    \label{tab:mnist_transfer_pgdlinf}
    \resizebox{1\columnwidth}{!}{%
    \begin{tabular}{|*{13}{c|}}
        \hline
        \multicolumn{13}{|c|}{\textbf{PGD($L_\infty$)}}\\
        \hline
        \multicolumn{13}{|l|}{\textbf{CCAT}}\\
        \hline
          $\tau$ & $\epsilon$ & \textbf{Correct} & \textbf{Rejected} & \textbf{Incorrect} & $\epsilon$ & \textbf{Correct} & \textbf{Rejected} & \textbf{Incorrect} & $\epsilon$ & \textbf{Correct} & \textbf{Rejected} & \textbf{Incorrect} \\
         \hline
          0.00 & 0.05 & 91.20 & 0.00   & \textbf{8.80} & 0.10 & 73.60 & 0.00   & \textbf{26.40} & 0.3 & 10.40 & 0.00   & \textbf{89.60}  \\
          0.30 &      & 0.00  & 100.00 & \textbf{0.00} &      & 0.00  & 100.00 & \textbf{0.00}  &     & 0.00  & 100.00 & \textbf{0.00}   \\
          0.70 &      & 0.00  & 100.00 & \textbf{0.00} &      & 0.00  & 100.00 & \textbf{0.00}  &     & 0.00  & 100.00 & \textbf{0.00}   \\
          1.00 &      & 0.00  & 100.00 & \textbf{0.00} &      & 0.00  & 100.00 & \textbf{0.00}  &     & 0.00  & 100.00 & \textbf{0.00}   \\
          \hline
          \multicolumn{13}{|l|}{\textbf{Ensemble}}\\
          \hline
          $\sigma$ & $\epsilon$ & \textbf{Correct} & \textbf{Rejected} & \textbf{Incorrect} & $\epsilon$ & \textbf{Correct} & \textbf{Rejected} & \textbf{Incorrect} & $\epsilon$ & \textbf{Correct} & \textbf{Rejected} & \textbf{Incorrect} \\
          \hline
          0.00 & 0.05 & 96.40 & 0.00  & \textbf{3.60} & 0.10 & 84.30 & 0.00  & \textbf{15.70} & 0.3 & 0.30 & 0.00 & \textbf{99.70}  \\ 
          0.30 &      & 96.40 & 0.00  & \textbf{3.60} &      & 84.30 & 0.00  & \textbf{15.70} &     & 0.30 & 0.00 & \textbf{99.70}  \\ 
          0.70 &      & 95.50 & 1.80  & \textbf{2.70} &      & 81.00 & 6.60  & \textbf{12.40} &     & 0.10 & 0.90 & \textbf{99.00}  \\ 
          1.00 &      & 88.40 & 10.70 & \textbf{0.90} &      & 62.80 & 32.10 & \textbf{5.10}  &     & 0.00 & 7.50 & \textbf{92.50}  \\
          \hline
          \multicolumn{13}{|l|}{\textbf{RMAggNet}}\\
          \hline
          EC & $\epsilon$ & \textbf{Correct} & \textbf{Rejected} & \textbf{Incorrect} & $\epsilon$ & \textbf{Correct} & \textbf{Rejected} & \textbf{Incorrect} & $\epsilon$ & \textbf{Correct} & \textbf{Rejected} & \textbf{Incorrect} \\
          \hline
          0 & 0.05 & 90.70 & 9.30 & \textbf{0.00} & 0.10 & 80.50 & 18.50 & \textbf{1.00} & 0.3 & 0.60 & 55.90 & \textbf{43.50}  \\ 
          1 &      & 94.70 & 4.90 & \textbf{0.40} &      & 88.00 & 10.30 & \textbf{1.70} &     & 1.60 & 35.40 & \textbf{63.00}  \\ 
          2 &      & 96.30 & 2.80 & \textbf{0.90} &      & 91.20 & 6.30  & \textbf{2.50} &     & 3.60 & 18.70 & \textbf{77.70}  \\ 
          3 &      & 97.60 & 0.80 & \textbf{1.60} &      & 93.40 & 2.50  & \textbf{4.10} &     & 5.60 & 7.30  & \textbf{87.10}  \\ 
          \hline
    \end{tabular}%
    }
    
    \end{subtable}
    
\end{table}

The results of the PGD $L_2$ transfer attack are in table \ref{tab:mnist_transfer_pgdl2}. We, again, aim to minimise incorrectness. We see that CCAT rejects nearly all of the inputs over all values of $\epsilon$, leading to 0 incorrect classifications. This is in contrast to both Ensemble and RMAggNet which are able to classify many of the inputs correctly at low $\epsilon$ values (0.5 and 1.00) with RMAggNet achieving approximately half the number of incorrect classifications, and even reducing to 0 for $\epsilon=0.50$. At higher $\epsilon$ we see a decrease in classification ability with both methods reporting high incorrectness values. However, RMAggNet is still able to approximately halve the number of incorrect classifications compared to Ensemble.

\begin{table}[htbp]
    \caption{Results for the transfer attacks using PGD $L_2$. Table \ref{tab:mnist_surrogate_pgdl2_res} shows the accuracy of the surrogate model on the PGD $L_2$ adversarial datasets. Table \ref{tab:mnist_transfer_pgdl2} shows the results of the adversarial datasets on the CCAT, Ensemble and RMAggNet models.}
    \centering
    \begin{subtable}[h]{1\textwidth}
        \caption{Accuracy of the surrogate MNIST classifier on the adversarial datasets generated using PGD $L_2$ with different perturbation budgets ($\epsilon$).}
        \label{tab:mnist_surrogate_pgdl2_res}
        \begin{center}
            \begin{tabular}{|c|c|}
                \hline
                 $\epsilon$ & \textbf{Accuracy (\%)} \\
                 \hline
                 0.00 & 99.10 \\
                 0.50 & 97.00 \\
                 1.00 & 87.00 \\
                 3.00 &  7.40 \\
                 \hline
            \end{tabular}
        \end{center}
    \end{subtable}
    \begin{subtable}[h]{1\columnwidth}
    \caption{Percentage of correct, rejected and incorrect classifications of the models using transfer attacks on a surrogate MNIST classifier using the PGD $L_2$ attack. Lower incorrectness is better.}
    \label{tab:mnist_transfer_pgdl2}
    \resizebox{1\columnwidth}{!}{
    \begin{tabular}{|*{13}{c|}}
        \hline
        \multicolumn{13}{|c|}{\textbf{PGD($L_2$)}}\\
        \hline
        \multicolumn{13}{|l|}{\textbf{CCAT}}\\
        \hline
          $\tau$ & $\epsilon$ & \textbf{Correct} & \textbf{Rejected} & \textbf{Incorrect} & $\epsilon$ & \textbf{Correct} & \textbf{Rejected} & \textbf{Incorrect} & $\epsilon$ & \textbf{Correct} & \textbf{Rejected} & \textbf{Incorrect} \\
         \hline
          0.00 & 0.50 & 98.20 & 0.00   & \textbf{1.80} & 1.00 & 93.70 & 0.00   & \textbf{6.30} & 3.00 & 32.80 & 0.00   & \textbf{67.20}  \\
          0.30 &      & 9.80  & 90.20  & \textbf{0.00} &      & 0.00  & 100.00 & \textbf{0.00} &      & 0.00  & 100.00 & \textbf{0.00}   \\
          0.70 &      & 1.50  & 98.50  & \textbf{0.00} &      & 0.00  & 100.00 & \textbf{0.00} &      & 0.00  & 100.00 & \textbf{0.00}   \\
          1.00 &      & 0.00  & 100.00 & \textbf{0.00} &      & 0.00  & 100.00 & \textbf{0.00} &      & 0.00  & 100.00 & \textbf{0.00}   \\
          \hline
          \multicolumn{13}{|l|}{\textbf{Ensemble}}\\
          \hline
          $\sigma$ & $\epsilon$ & \textbf{Correct} & \textbf{Rejected} & \textbf{Incorrect} & $\epsilon$ & \textbf{Correct} & \textbf{Rejected} & \textbf{Incorrect} & $\epsilon$ & \textbf{Correct} & \textbf{Rejected} & \textbf{Incorrect} \\
          \hline
          0.00 & 0.50 & 97.40 & 0.00 & \textbf{2.60} & 1.00 & 92.90 & 0.00  & \textbf{7.10} & 3.00 & 16.60 & 0.00  & \textbf{83.40} \\ 
          0.30 &      & 97.40 & 0.00 & \textbf{2.60} &      & 92.90 & 0.00  & \textbf{7.10} &      & 16.60 & 0.00  & \textbf{83.40} \\ 
          0.70 &      & 97.00 & 1.30 & \textbf{1.70} &      & 90.80 & 3.80  & \textbf{5.40} &      & 13.10 & 6.60  & \textbf{80.30} \\
          1.00 &      & 91.70 & 7.60 & \textbf{0.70} &      & 79.90 & 17.90 & \textbf{2.20} &      & 8.60  & 27.90 & \textbf{63.50} \\
          \hline
          \multicolumn{13}{|l|}{\textbf{RMAggNet}}\\
          \hline
          EC & $\epsilon$ & \textbf{Correct} & \textbf{Rejected} & \textbf{Incorrect} & $\epsilon$ & \textbf{Correct} & \textbf{Rejected} & \textbf{Incorrect} & $\epsilon$ & \textbf{Correct} & \textbf{Rejected} & \textbf{Incorrect} \\
          \hline
          0 & 0.50 & 92.00 & 8.00 & \textbf{0.00} & 1.00 & 86.30 & 13.00 & \textbf{0.70} & 3.00 & 18.00 & 62.90 & \textbf{19.10}  \\ 
          1 &      & 95.50 & 4.20 & \textbf{0.30} &      & 90.70 & 8.20  & \textbf{1.10} &      & 29.80 & 39.50 & \textbf{30.70}  \\ 
          2 &      & 97.10 & 2.10 & \textbf{0.80} &      & 93.40 & 4.60  & \textbf{2.00} &      & 35.70 & 24.40 & \textbf{39.90}  \\ 
          3 &      & 97.80 & 0.70 & \textbf{1.50} &      & 95.80 & 1.50  & \textbf{2.70} &      & 44.20 & 8.70  & \textbf{47.10}  \\ 
          \hline
    \end{tabular}
    }
    \end{subtable}
    
\end{table}

Table \ref{tab:mnist_direct_pgdlinf} shows the results of open-box PGD $L_\infty$ adversarial attacks where the adversarial perturbations are generated using the models themselves rather than a surrogate. Details on how these adversarial images were generated can be found in Section \ref{sec:adv_gen}. The aim here is to minimise incorrectness. These attacks are slightly more effective than the transfer attacks. They follow a similar pattern to the transfer attacks, with RMAggNet allowing adversarial examples to be corrected leading to higher correctness at low values of $\epsilon$, which becomes less effective at higher $\epsilon$. CCAT is able to achieve 0\% incorrectness by rejecting all adversarial examples at $\tau > 0$. This is an advantage at $\epsilon = 0.30$, since both Ensemble and RMAggNet have high number of incorrect classifications, with RMAggNet having a minimum incorrectness of 41\% compared to the Ensemble method at 94\%. However, at the lower perturbation values RMAggNet is able to correctly classify a majority of the adversarial examples with fairly small amounts of incorrect classifications whereas CCAT rejects all inputs. RMAggNet is also able to outperform the Ensemble method at these low perturbation values as well, with similar correctness at $\epsilon=0.05$, but with much lower incorrectness, and a much higher correctness at $\epsilon=0.10$, with significantly lower incorrectness.

\begin{table}[htbp]
    \caption{Results for open-box attacks using PGD $L_\infty$ on the MNIST dataset at different perturbation budgets ($\epsilon$). Lower incorrectness is better.}
    \label{tab:mnist_direct_pgdlinf}
    \centering
    \resizebox{1\columnwidth}{!}{%
    \begin{tabular}{|*{13}{c|}}
        \hline
        \multicolumn{13}{|c|}{\textbf{PGD($L_\infty$)}}\\
        \hline
        \multicolumn{13}{|l|}{\textbf{CCAT}}\\
        \hline
          $\tau$ & $\epsilon$ & \textbf{Correct} & \textbf{Rejected} & \textbf{Incorrect} & $\epsilon$ & \textbf{Correct} & \textbf{Rejected} & \textbf{Incorrect} & $\epsilon$ & \textbf{Correct} & \textbf{Rejected} & \textbf{Incorrect} \\
         \hline
          0.00 & 0.05 & 44.80 & 0.00   & \textbf{55.20} & 0.10 & 16.60 & 0.00   & \textbf{83.40} & 0.3 & 0.30 & 0.00   & \textbf{99.70}  \\
          0.30 &      & 0.00  & 100.00 & \textbf{0.00}  &      & 0.00  & 100.00 & \textbf{0.00}  &     & 0.00 & 100.00 & \textbf{0.00}   \\
          0.70 &      & 0.00  & 100.00 & \textbf{0.00}  &      & 0.00  & 100.00 & \textbf{0.00}  &     & 0.00 & 100.00 & \textbf{0.00}   \\
          1.00 &      & 0.00  & 100.00 & \textbf{0.00}  &      & 0.00  & 100.00 & \textbf{0.00}  &     & 0.00 & 100.00 & \textbf{0.00}   \\
          \hline
          \multicolumn{13}{|l|}{\textbf{Ensemble}}\\
          \hline
          $\sigma$ & $\epsilon$ & \textbf{Correct} & \textbf{Rejected} & \textbf{Incorrect} & $\epsilon$ & \textbf{Correct} & \textbf{Rejected} & \textbf{Incorrect} & $\epsilon$ & \textbf{Correct} & \textbf{Rejected} & \textbf{Incorrect} \\
          \hline
          0.00 & 0.05 & 94.80 & 0.00  & \textbf{5.20} & 0.10 & 76.90 & 0.00  & \textbf{23.10} & 0.3 & 0.00 & 0.00 & \textbf{100.00}  \\ 
          0.30 &      & 94.80 & 0.00  & \textbf{5.20} &      & 76.90 & 0.00  & \textbf{23.10} &     & 0.00 & 0.00 & \textbf{100.00}  \\ 
          0.70 &      & 92.60 & 3.10  & \textbf{4.30} &      & 72.50 & 7.20  & \textbf{20.30} &     & 0.00 & 0.50 & \textbf{99.50}  \\ 
          1.00 &      & 85.60 & 12.90 & \textbf{1.50} &      & 50.60 & 39.80 & \textbf{9.60}  &     & 0.00 & 5.50 & \textbf{94.50}  \\
          \hline
          \multicolumn{13}{|l|}{\textbf{RMAggNet}}\\
          \hline
          EC & $\epsilon$ & \textbf{Correct} & \textbf{Rejected} & \textbf{Incorrect} & $\epsilon$ & \textbf{Correct} & \textbf{Rejected} & \textbf{Incorrect} & $\epsilon$ & \textbf{Correct} & \textbf{Rejected} & \textbf{Incorrect} \\
          \hline
          0 & 0.05 & 79.00 & 20.40 & \textbf{0.60} & 0.10 & 41.40 & 56.40 & \textbf{2.20} & 0.3 & 0.00 & 58.90 & \textbf{41.10}  \\ 
          1 &      & 90.40 & 8.30  & \textbf{1.30} &      & 74.80 & 21.40 & \textbf{3.80} &     & 0.10 & 35.50 & \textbf{64.40}  \\ 
          2 &      & 94.00 & 4.00  & \textbf{2.00} &      & 84.50 & 10.50 & \textbf{5.00} &     & 0.70 & 20.50 & \textbf{78.80}  \\ 
          3 &      & 95.70 & 1.50  & \textbf{2.80} &      & 89.10 & 3.60  & \textbf{7.30} &     & 2.10 & 9.00  & \textbf{88.90}  \\ 
          \hline
    \end{tabular}%
    }
    
\end{table}

The PGD $L_2$ open-box attack (table \ref{tab:mnist_direct_pgdl2}) mirrors results from the closed-box transfer attack (table \ref{tab:mnist_transfer_pgdl2}). CCAT is able to achieve the lowest incorrectness results at 0\% for all $\tau > 0$ over all $\epsilon$. However, these low incorrectness values are due to rejection in situations where some amount of inputs could be correctly classified. Comparing Ensemble and RMAggNet, RMAggNet is able to achieve consistently lower incorrectness scores, with equal or higher correctness than Ensemble. At $\epsilon=3.0$, the incorrectness of both Ensemble and RMAggNet increase considerably, showing that CCAT is the preferable model if we expect strong adversaries to be present.

\begin{table}[htbp]
    \caption{Results for open-box attacks using PGD $L_2$ on the MNIST dataset at different perturbation budgets ($\epsilon$). Lower incorrectness is better.}
    \label{tab:mnist_direct_pgdl2}
    \centering
    \resizebox{1\columnwidth}{!}{
    \begin{tabular}{|*{13}{c|}}
        \hline
        \multicolumn{13}{|c|}{\textbf{PGD($L_2$)}}\\
        \hline
        \multicolumn{13}{|l|}{\textbf{CCAT}}\\
        \hline
          $\tau$ & $\epsilon$ & \textbf{Correct} & \textbf{Rejected} & \textbf{Incorrect} & $\epsilon$ & \textbf{Correct} & \textbf{Rejected} & \textbf{Incorrect} & $\epsilon$ & \textbf{Correct} & \textbf{Rejected} & \textbf{Incorrect} \\
         \hline
          0.00 & 0.50 & 75.60 & 0.00   & \textbf{24.40} & 1.00 & 42.90 & 0.00   & \textbf{57.10} & 3.00 & 3.70 & 0.00   & \textbf{96.30}  \\
          0.30 &      & 0.00  & 100.00 & \textbf{0.00}  &      & 0.00  & 100.00 & \textbf{0.00}  &      & 0.00 & 100.00 & \textbf{0.00}   \\
          0.70 &      & 0.00  & 100.00 & \textbf{0.00}  &      & 0.00  & 100.00 & \textbf{0.00}  &      & 0.00 & 100.00 & \textbf{0.00}   \\
          1.00 &      & 0.00  & 100.00 & \textbf{0.00}  &      & 0.00  & 100.00 & \textbf{0.00}  &      & 0.00 & 100.00 & \textbf{0.00}   \\
          \hline
          \multicolumn{13}{|l|}{\textbf{Ensemble}}\\
          \hline
          $\sigma$ & $\epsilon$ & \textbf{Correct} & \textbf{Rejected} & \textbf{Incorrect} & $\epsilon$ & \textbf{Correct} & \textbf{Rejected} & \textbf{Incorrect} & $\epsilon$ & \textbf{Correct} & \textbf{Rejected} & \textbf{Incorrect} \\
          \hline
          0.00 & 0.50 & 97.00 & 0.00 & \textbf{3.00} & 1.00 & 89.60 & 0.00  & \textbf{10.40} & 3.00 & 51.60 & 0.00 & \textbf{48.40} \\ 
          0.30 &      & 97.00 & 0.00 & \textbf{3.00} &      & 89.60 & 0.00  & \textbf{10.40} &      & 51.60 & 0.00 & \textbf{48.40} \\ 
          0.70 &      & 96.30 & 1.40 & \textbf{2.30} &      & 87.80 & 4.40  & \textbf{7.80}  &      & 51.40 & 0.60 & \textbf{48.00} \\
          1.00 &      & 90.20 & 9.00 & \textbf{0.80} &      & 79.70 & 17.10 & \textbf{3.20}  &      & 50.90 & 3.00 & \textbf{46.10} \\
          \hline
          \multicolumn{13}{|l|}{\textbf{RMAggNet}}\\
          \hline
          EC & $\epsilon$ & \textbf{Correct} & \textbf{Rejected} & \textbf{Incorrect} & $\epsilon$ & \textbf{Correct} & \textbf{Rejected} & \textbf{Incorrect} & $\epsilon$ & \textbf{Correct} & \textbf{Rejected} & \textbf{Incorrect} \\
          \hline
          0 & 0.50 & 85.50 & 14.30 & \textbf{0.20} & 1.00 & 75.40 & 23.10 & \textbf{1.50} & 3.00 & 46.50 & 34.10 & \textbf{19.40}  \\ 
          1 &      & 93.40 & 5.80  & \textbf{0.80} &      & 86.00 & 11.50 & \textbf{2.50} &      & 49.00 & 22.70 & \textbf{28.30}  \\ 
          2 &      & 95.80 & 2.80  & \textbf{1.40} &      & 90.60 & 6.20  & \textbf{3.20} &      & 51.70 & 14.10 & \textbf{34.20}  \\ 
          3 &      & 96.90 & 0.90  & \textbf{2.20} &      & 92.80 & 3.10  & \textbf{4.10} &      & 54.80 & 7.00  & \textbf{38.20}  \\ 
          \hline
    \end{tabular}
    }
    
\end{table}

Our final attack on the MNIST dataset is the Boundary ($L_2$) attack. The results are in tables \ref{tab:mnist_transfer_boundaryl2} and \ref{tab:mnist_direct_boundaryl2}, where we look to minimise incorrectness. Looking at the transfer attack in table \ref{tab:mnist_transfer_boundaryl2}, we see that CCAT performs worse compared to the previous attacks, with a rejection rate between 98-100\% only achieving the 100\% mark at the strictest confidence threshold. The results on the Ensemble and RMAggNet methods are similar to the previous adversarial attacks, where the percentage of correctly classified inputs are similar for low $\epsilon$, with RMAggNet able to achieve approximately half the number of incorrect classifications (dropping to 0 for $\epsilon=1.50$ for $EC \leq 2$). However, at higher $\epsilon$ we see both Ensemble and RMAggNet retain their high correctness and a low number of incorrect classifications, with RMAggNet reporting 0\% incorrectness in many cases. This can also be seen in CCAT as the low thresholds ($\tau = 0$) show over 85\% correctness. Looking at the effect that the Boundary attack has on the surrogate model (table \ref{tab:mnist_surrogate_res_boundary}) it indicates that the adversarial examples produced for it focus on making subtle changes to features which are unique to the surrogate model and do not generalise to the CWR methods.

Table \ref{tab:mnist_direct_boundaryl2} shows the power of the Boundary ($L_2$) attack in a open-box setting. Both Ensemble and RMAggNet show reduced correctness over all $\epsilon$, which becomes more pronounced as $\epsilon$ increases. These adversarial examples appear to affect RMAggNet to a greater degree leading to a 10-30\% reduction in correctness compared to Ensemble. However, RMAggNet is able to achieve a significantly lower number of incorrect classifications compared to Ensemble, leveraging the reject option effectively to reduce the incorrectness to 0\% at $EC \leq 1$ for all $\epsilon$. It is worth noting that no open-box attacks could be carried out on CCAT since no initial adversarial examples could be found by the Boundary method operating at the same conditions as Ensemble and RMAggNet. 

\begin{table}[htbp]
    \caption{Results for the transfer attacks using Boundary ($L_2$). Table \ref{tab:mnist_surrogate_res_boundary} shows the accuracy of the surrogate model on the Boundary ($L_2$) adversarial datasets. Table \ref{tab:mnist_transfer_boundaryl2} shows the results of the adversarial datasets on the CCAT, Ensemble and RMAggNet models.}
    \centering
    \begin{subtable}[h]{1\columnwidth}
        \caption{Accuracy of the surrogate MNIST classifier on the adversarial datasets generated using the Boundary ($L_2$) attack with different perturbation budgets ($\epsilon$).}
        \label{tab:mnist_surrogate_res_boundary}
        \begin{center}
            \begin{tabular}{|c|c|}
                \hline
                 $\epsilon$ & \textbf{Accuracy (\%)} \\
                 \hline
                 0.00 & 99.10 \\
                 1.50 & 62.00 \\
                 2.00 & 28.30 \\
                 3.00 & 0.90  \\
                 \hline
            \end{tabular}%
        \end{center}
    \end{subtable}
    \begin{subtable}[h]{\textwidth}
    \caption{Percentage of correct, rejected and incorrect classifications of the models using transfer attacks on a surrogate MNIST classifier using the Boundary ($L_2$) attack. Lower incorrectness is better.}
    \label{tab:mnist_transfer_boundaryl2}
    \resizebox{1\columnwidth}{!}{%
    \begin{tabular}{|*{13}{c|}}
        \hline
        \multicolumn{13}{|c|}{\textbf{Boundary ($L_2$)}}\\
        \hline
        \multicolumn{13}{|l|}{\textbf{CCAT}}\\
        \hline
          $\tau$ & $\epsilon$ & \textbf{Correct} & \textbf{Rejected} & \textbf{Incorrect} & $\epsilon$ & \textbf{Correct} & \textbf{Rejected} & \textbf{Incorrect} & $\epsilon$ & \textbf{Correct} & \textbf{Rejected} & \textbf{Incorrect} \\
         \hline
          0.00 & 1.50 & 94.00 & 0.00   & \textbf{6.00} & 2.00 & 90.60 & 0.00   & \textbf{9.40} & 3.00 & 86.70 & 0.00   & \textbf{13.30}  \\
          0.30 &      & 0.60  & 98.90  & \textbf{0.50} &      & 0.60  & 98.90  & \textbf{0.50} &      & 0.60  & 98.90  & \textbf{0.50}   \\
          0.70 &      & 0.20  & 99.60  & \textbf{0.20} &      & 0.20  & 99.60  & \textbf{0.20} &      & 0.20  & 99.60  & \textbf{0.20}   \\
          1.00 &      & 0.00  & 100.00 & \textbf{0.00} &      & 0.00  & 100.00 & \textbf{0.00} &      & 0.00  & 100.00 & \textbf{0.00}   \\
          \hline
          \multicolumn{13}{|l|}{\textbf{Ensemble}}\\
          \hline
          $\sigma$ & $\epsilon$ & \textbf{Correct} & \textbf{Rejected} & \textbf{Incorrect} & $\epsilon$ & \textbf{Correct} & \textbf{Rejected} & \textbf{Incorrect} & $\epsilon$ & \textbf{Correct} & \textbf{Rejected} & \textbf{Incorrect} \\
          \hline
          0.00 & 1.50 & 97.70 & 0.00  & \textbf{2.30} & 2.00 & 97.30 & 0.00  & \textbf{2.70} & 3.00 & 97.10 & 0.00  & \textbf{2.90}  \\
          0.30 &      & 97.70 & 0.00  & \textbf{2.30} &      & 97.30 & 0.00  & \textbf{2.70} &      & 97.10 & 0.00  & \textbf{2.90}  \\
          0.70 &      & 93.10 & 5.70  & \textbf{1.20} &      & 91.40 & 7.40  & \textbf{1.20} &      & 90.40 & 8.40  & \textbf{1.20}  \\
          1.00 &      & 68.10 & 31.80 & \textbf{0.10} &      & 55.00 & 44.90 & \textbf{0.10} &      & 48.10 & 51.80 & \textbf{0.10}  \\
          \hline
          \multicolumn{13}{|l|}{\textbf{RMAggNet}}\\
          \hline
          EC & $\epsilon$ & \textbf{Correct} & \textbf{Rejected} & \textbf{Incorrect} & $\epsilon$ & \textbf{Correct} & \textbf{Rejected} & \textbf{Incorrect} & $\epsilon$ & \textbf{Correct} & \textbf{Rejected} & \textbf{Incorrect} \\
          \hline
          0 & 1.50 & 81.60 & 18.40 & \textbf{0.00} & 2.00 & 77.10 & 22.90 & \textbf{0.00} & 3.00 & 74.00 & 26.00 & \textbf{0.00}  \\ 
          1 &      & 91.50 & 8.50  & \textbf{0.00} &      & 90.00 & 10.00 & \textbf{0.00} &      & 89.20 & 10.80 & \textbf{0.00}  \\ 
          2 &      & 95.90 & 4.10  & \textbf{0.00} &      & 95.10 & 4.90  & \textbf{0.00} &      & 94.80 & 5.20  & \textbf{0.00}  \\ 
          3 &      & 97.60 & 1.70  & \textbf{0.70} &      & 97.40 & 1.90  & \textbf{0.70} &      & 97.40 & 1.90  & \textbf{0.70}  \\ 
          \hline
    \end{tabular}%
    }
    
    \end{subtable}
    
\end{table}

\begin{table}[htbp]
    \caption{Results for direct attacks using Boundary ($L_2$) on the MNIST dataset at different perturbation budgets ($\epsilon$). Lower incorrectness is better.}
    \label{tab:mnist_direct_boundaryl2}
    \centering
    \resizebox{1\columnwidth}{!}{%
    \begin{tabular}{|*{13}{c|}}
        \hline
        \multicolumn{13}{|c|}{\textbf{Boundary ($L_2$)}}\\
        \hline
        \multicolumn{13}{|l|}{\textbf{CCAT}}\\
        \hline
          $\tau$ & $\epsilon$ & \textbf{Correct} & \textbf{Rejected} & \textbf{Incorrect} & $\epsilon$ & \textbf{Correct} & \textbf{Rejected} & \textbf{Incorrect} & $\epsilon$ & \textbf{Correct} & \textbf{Rejected} & \textbf{Incorrect} \\
         \hline
          \multicolumn{13}{|c|}{Failed to find initial adversaries}\\
          \hline
          \multicolumn{13}{|l|}{\textbf{Ensemble}}\\
          \hline
          $\sigma$ & $\epsilon$ & \textbf{Correct} & \textbf{Rejected} & \textbf{Incorrect} & $\epsilon$ & \textbf{Correct} & \textbf{Rejected} & \textbf{Incorrect} & $\epsilon$ & \textbf{Correct} & \textbf{Rejected} & \textbf{Incorrect} \\
          \hline
          0.00 & 1.50 & 81.90 & 0.00  & \textbf{18.10} & 2.00 & 64.50 & 0.00  & \textbf{35.50} & 3.00 & 47.30 & 0.00  & \textbf{52.70} \\
          0.30 &      & 81.90 & 0.00  & \textbf{18.10} &      & 64.50 & 0.00  & \textbf{35.50} &      & 47.30 & 0.00  & \textbf{52.70} \\
          0.70 &      & 64.80 & 33.80 & \textbf{1.40}  &      & 32.40 & 65.90 & \textbf{1.70}  &      & 2.70  & 95.10 & \textbf{2.20}  \\
          1.00 &      & 45.00 & 55.00 & \textbf{0.00}  &      & 17.50 & 82.50 & \textbf{0.00}  &      & 0.80  & 99.20 & \textbf{0.00}  \\
          \hline
          \multicolumn{13}{|l|}{\textbf{RMAggNet}}\\
          \hline
          EC & $\epsilon$ & \textbf{Correct} & \textbf{Rejected} & \textbf{Incorrect} & $\epsilon$ & \textbf{Correct} & \textbf{Rejected} & \textbf{Incorrect} & $\epsilon$ & \textbf{Correct} & \textbf{Rejected} & \textbf{Incorrect} \\
          \hline
          0 & 1.50 & 53.30 & 46.70 & \textbf{0.00} & 2.00 & 22.30 & 77.70 & \textbf{0.00} & 3.00 & 0.70  & 99.30 & \textbf{0.00}  \\
          1 &      & 59.20 & 40.80 & \textbf{0.00} &      & 25.20 & 74.80 & \textbf{0.00} &      & 0.70  & 99.30 & \textbf{0.00}  \\
          2 &      & 61.70 & 38.20 & \textbf{0.10} &      & 26.00 & 73.90 & \textbf{0.10} &      & 0.90  & 99.00 & \textbf{0.10}  \\
          3 &      & 70.70 & 24.50 & \textbf{4.80} &      & 41.60 & 49.70 & \textbf{8.70} &      & 20.00 & 67.70 & \textbf{12.30} \\
          
          \hline
    \end{tabular}%
    }
\end{table}

Images corresponding to the PGD $L_\infty$ attacks can be found in figure \ref{fig:MNIST_PGDLinf_images}, with PGD $L_2$ in \ref{fig:MNIST_PGDL2_images} and Boundary in \ref{fig:MNIST_boundary_images}.

\begin{figure}
    \begin{subfigure}{0.3\textwidth}
    
    \tikzset{every picture/.style={line width=0.75pt}}
        \begin{tikzpicture}[x=0.75pt,y=0.75pt,yscale=-1,xscale=1]
        
        \draw (-45,0) node [anchor=north west][inner sep=0.75pt]    {Surrogate Model};
        \draw (0,30) node  {\includegraphics[width=1.75\linewidth]{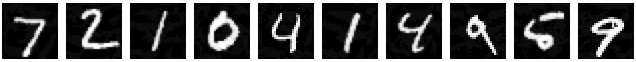}};
        \draw (0,60) node  {\includegraphics[width=1.75\linewidth]{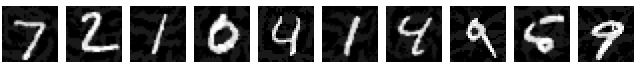}};
        \draw (0,90) node  {\includegraphics[width=1.75\linewidth]{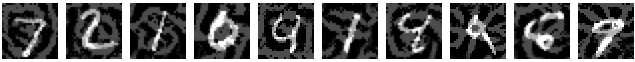}};
        
        \draw (255,0) node [anchor=north west][inner sep=0.75pt]    {CCAT Model};
        \draw (300,30) node  {\includegraphics[width=1.75\linewidth]{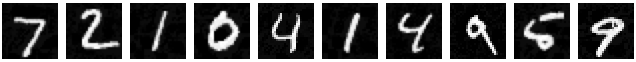}};
        \draw (300,60) node  {\includegraphics[width=1.75\linewidth]{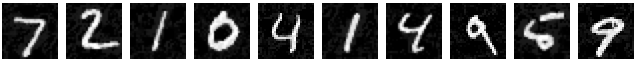}};
        \draw (300,90) node  {\includegraphics[width=1.75\linewidth]{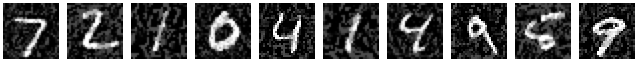}};

        \draw (-45,120) node [anchor=north west][inner sep=0.75pt]    {Ensemble Model};
        \draw (0,150) node  {\includegraphics[width=1.75\linewidth]{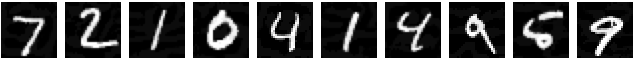}};
        \draw (0,180) node  {\includegraphics[width=1.75\linewidth]{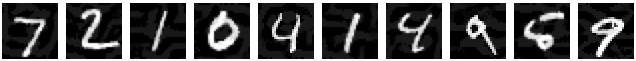}};
        \draw (0,210) node  {\includegraphics[width=1.75\linewidth]{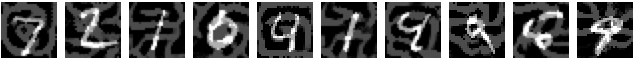}};

        \draw (255,120) node [anchor=north west][inner sep=0.75pt]    {RMAggNet Model};
        \draw (300,150) node  {\includegraphics[width=1.75\linewidth]{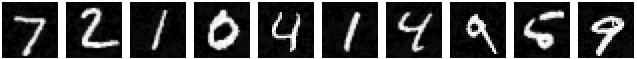}};
        \draw (300,180) node  {\includegraphics[width=1.75\linewidth]{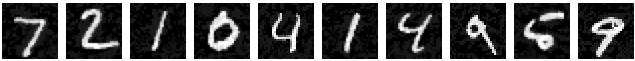}};
        \draw (300,210) node  {\includegraphics[width=1.75\linewidth]{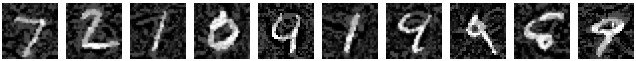}};
        
        \end{tikzpicture}
    \end{subfigure}
    \caption{Adversarial images generated using the PGD $L_\infty$ attack on the MNIST dataset across all models. Each architecture shows three rows of images, where each row shows the adversarial images generated at a different $\epsilon \in \{0.05, 0.10, 0.30\}$.}
    \label{fig:MNIST_PGDLinf_images}
\end{figure}

\begin{figure}

    \begin{subfigure}{0.3\columnwidth}
    \centering
    \tikzset{every picture/.style={line width=0.75pt}}
        \begin{tikzpicture}[x=0.75pt,y=0.75pt,yscale=-1,xscale=1]
        
        \draw (-45,0) node [anchor=north west][inner sep=0.75pt]    {Surrogate Model};
        \draw (0,30) node  {\includegraphics[width=1.75\linewidth]{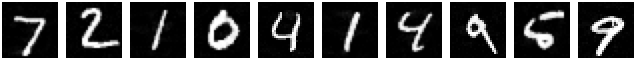}};
        \draw (0,60) node  {\includegraphics[width=1.75\linewidth]{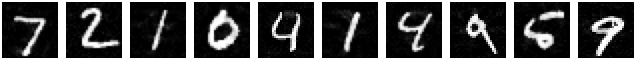}};
        \draw (0,90) node  {\includegraphics[width=1.75\linewidth]{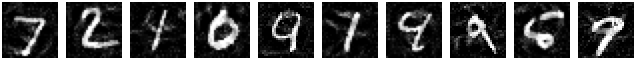}};

        \draw (255,0) node [anchor=north west][inner sep=0.75pt]    {CCAT Model};
        \draw (300,30) node  {\includegraphics[width=1.75\linewidth]{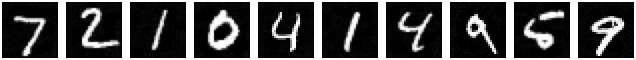}};
        \draw (300,60) node  {\includegraphics[width=1.75\linewidth]{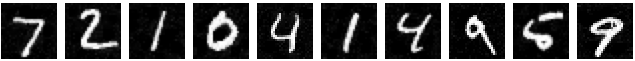}};
        \draw (300,90) node  {\includegraphics[width=1.75\linewidth]{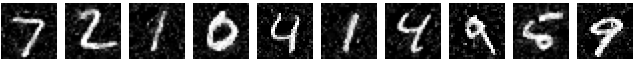}};
        
        \draw (-45,120) node [anchor=north west][inner sep=0.75pt]    {Ensemble Model};
        \draw (0,150) node  {\includegraphics[width=1.75\linewidth]{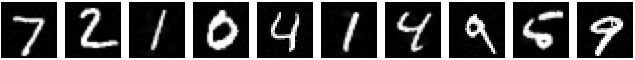}};
        \draw (0,180) node  {\includegraphics[width=1.75\linewidth]{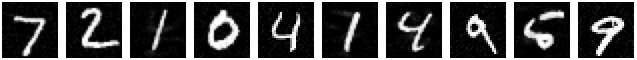}};
        \draw (0,210) node  {\includegraphics[width=1.75\linewidth]{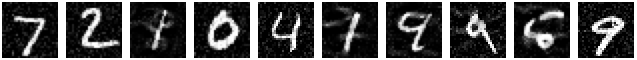}};

        \draw (255,120) node [anchor=north west][inner sep=0.75pt]    {RMAggNet Model};
        \draw (300,150) node  {\includegraphics[width=1.75\linewidth]{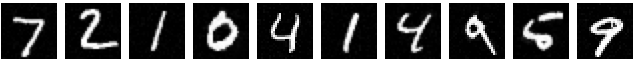}};
        \draw (300,180) node  {\includegraphics[width=1.75\linewidth]{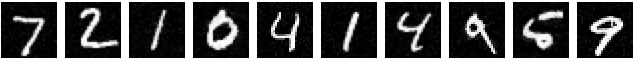}};
        \draw (300,210) node  {\includegraphics[width=1.75\linewidth]{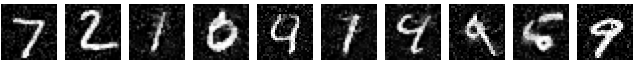}};

        \end{tikzpicture}
    
    \end{subfigure}
    \caption{Adversarial images generated using the PGD $L_2$ attack on the MNIST dataset across all models. Each architecture shows three rows of images, where each row shows the adversarial images generated at a different $\epsilon \in \{0.5, 1.0, 3.0\}$.}
    \label{fig:MNIST_PGDL2_images}
\end{figure}

\begin{figure}
    \begin{subfigure}{0.3\linewidth}
    \centering
    \tikzset{every picture/.style={line width=0.75pt}}
        \begin{tikzpicture}[x=0.75pt,y=0.75pt,yscale=-1,xscale=1]
        
        \draw (-45,0) node [anchor=north west][inner sep=0.75pt]    {Surrogate Model};
        \draw (0,30) node  {\includegraphics[width=1.75\linewidth]{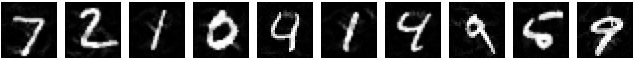}};
        \draw (0,60) node  {\includegraphics[width=1.75\linewidth]{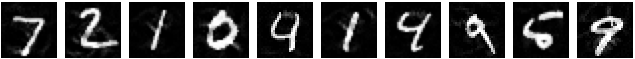}};
        \draw (0,90) node  {\includegraphics[width=1.75\linewidth]{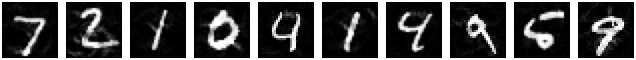}};

        \draw (255,0) node [anchor=north west][inner sep=0.75pt]    {Ensemble Model};
        \draw (300,30) node  {\includegraphics[width=1.75\linewidth]{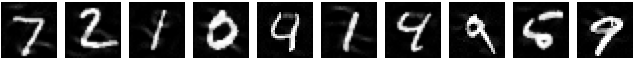}};
        \draw (300,60) node  {\includegraphics[width=1.75\linewidth]{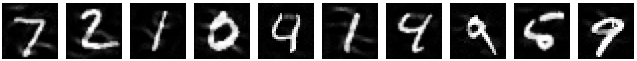}};
        \draw (300,90) node  {\includegraphics[width=1.75\linewidth]{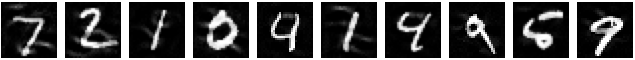}};

        \draw (100,120) node [anchor=north west][inner sep=0.75pt]    {RMAggNet Model};
        \draw (150,150) node  {\includegraphics[width=1.75\linewidth]{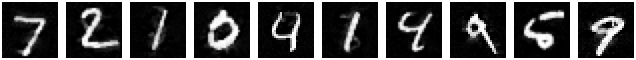}};
        \draw (150,180) node  {\includegraphics[width=1.75\linewidth]{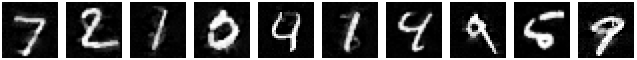}};
        \draw (150,210) node  {\includegraphics[width=1.75\linewidth]{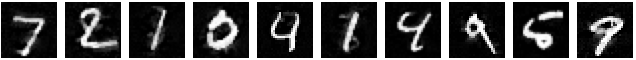}};
        
        \end{tikzpicture}
    \end{subfigure}
    \caption{Adversarial images generated using the Boundary $L_2$ attack on the MNIST dataset across all models. Each architecture shows four rows of images, where each row shows the adversarial images generated at a different $\epsilon \in \{1.5, 2.0, 3.0\}$. Note that CCAT does not have any adversarial images due to the Boundary method not being able to find any initial successful perturbations close enough to the classification boundary.}
    \label{fig:MNIST_boundary_images}
\end{figure}

}

\subsection{EMNIST Dataset} \label{sec:res_emnist}
Results for the EMNIST dataset use RMAggNet with $m=5$, $r=1$ which gives us 32 networks with 7 bits of error correction (EC). We also use 32 networks for the Ensemble method for parity. All methods use ResNet-18 models \cite{he2016deep}.
Table \ref{tab:emnist_res} shows the results on the clean EMNIST dataset where we expect to maximise correctness. All three models perform similarly, with Ensemble achieving the highest correctness, closely followed by RMAggNet and CCAT. However, all models come close to state-of-the-art performance (91.06\%) \cite{jeevan2022wavemix}, with minimal negative impacts from the adversarial defence.

\begin{table}[!htbp]
    \caption{Results for the clean EMNIST dataset showing the percentage of classifications that are correct, rejected and incorrect. Bold text indicates the metric of interest. Higher correctness is better.}

    \label{tab:emnist_res}
      
    \begin{subtable}[h]{0.32938\columnwidth}
    \resizebox{1\columnwidth}{!}{
        \begin{tabular}{|*{4}{c|}}
            \hline
            \multicolumn{4}{|l|}{\textbf{CCAT}}\\
            \hline
            \textbf{$\tau$} & \textbf{Correct} & \textbf{Rejected} & \textbf{Incorrect} \\
            \hline
            0    &  \textbf{88.68} &   0.00 & 11.32\\
            0.10 &  \textbf{88.60} &   0.15 & 11.24\\
            0.20 &  \textbf{87.54} &   2.41 & 10.05\\
            0.30 &  \textbf{85.46} &   6.45 &  8.09\\
            0.40 &  \textbf{83.16} &  10.29 &  6.55\\
            0.50 &  \textbf{80.70} &  14.22 &  5.08\\
            0.60 &  \textbf{77.95} &  18.03 &  4.02\\
            0.70 &  \textbf{74.23} &  22.71 &  3.06\\
            0.80 &  \textbf{69.48} &  28.46 &  2.06\\
            0.90 &  \textbf{60.74} &  38.05 &  1.20\\
            1.0  &  \textbf{ 0.00} & 100.00 &  0.00\\
            \hline
        \end{tabular}
        }
        \label{tab:emnist_ccat}
    \end{subtable}
    \begin{subtable}[h]{0.32938\columnwidth}
    \resizebox{1\columnwidth}{!}{
        \begin{tabular}{|*{4}{c|}}
            \hline
            \multicolumn{4}{|l|}{\textbf{Ensemble}}\\
            \hline
            \textbf{$\sigma$} & \textbf{Correct} & \textbf{Rejected} & \textbf{Incorrect} \\
            \hline
            0    & \textbf{89.78} &  0.00 & 10.22 \\
            0.10 & \textbf{89.78} &  0.00 & 10.22 \\
            0.20 & \textbf{89.78} &  0.01 & 10.21 \\
            0.30 & \textbf{89.76} &  0.05 & 10.19 \\
            0.40 & \textbf{89.70} &  0.28 & 10.03 \\
            0.50 & \textbf{89.20} &  1.41 &  9.39 \\
            0.60 & \textbf{87.76} &  4.45 &  7.79 \\
            0.70 & \textbf{85.94} &  7.68 &  6.38 \\
            0.80 & \textbf{83.89} & 11.05 &  5.06 \\
            0.90 & \textbf{80.60} & 15.60 &  3.80 \\
            1.0  & \textbf{68.34} & 30.15 &  1.51 \\
            \hline
        \end{tabular}
        }
        \label{tab:emnist_ensemble}
    \end{subtable}    
    \begin{subtable}[h]{0.32938\columnwidth}
        \resizebox{1\columnwidth}{!}{
        \begin{tabular}{|*{4}{c|}}
            \hline
            \multicolumn{4}{|l|}{\textbf{RMAggNet} $[32,6,16]_2$}\\
            \hline
            \textbf{EC} & \textbf{Correct} & \textbf{Rejected} & \textbf{Incorrect} \\
            \hline
            0 & \textbf{70.02} & 27.72 & 2.26 \\
            1 & \textbf{77.19} & 19.59 & 3.23 \\
            2 & \textbf{80.85} & 15.15 & 4.00 \\
            3 & \textbf{83.29} & 12.03 & 4.68 \\
            4 & \textbf{84.99} & 9.49 & 5.52 \\
            5 & \textbf{86.54} & 6.98 & 6.48 \\
            6 & \textbf{87.93} & 4.59 & 7.48 \\
            7 & \textbf{89.16} & 2.14 & 8.70 \\
            \hline
        \end{tabular}
        }
        \label{tab:emnist_rmaggnet}
    \end{subtable}
    
\end{table}

\longpaper{The results for the PGD $L_\infty$ transfer attacks are in table \ref{tab:emnist_transfer_pgdlinf} where we aim to minimise incorrectness, either through rejection or correct classifications. The CCAT model is able to reject all adversarial inputs for $\tau > 0$, however the $\tau = 0$ results provide us with some interesting insights. $\tau = 0$ effectively removes the confidence threshold which is used to reject inputs, this indicates that the CCAT model is strongly affected by the adversarial examples generated by the surrogate model. If we turn our attention to the Ensemble and RMAggNet methods, we see that RMAggNet is able to achieve a higher correctness than Ensemble with the difference becoming larger as $\epsilon$ increases. In addition, RMAggNet is able to achieve a similar minimum incorrectness with slightly higher correctness over all values of $\epsilon$.}{}

\longpaper{The results of the PGD $L_2$ transfer results are in table \ref{tab:emnist_transfer_pgdl2}. This shows that CCAT is, once again, able to reject all adversarial data for $\tau > 0$, resulting in 0 incorrect classification across all $\epsilon$ values. We can also see that RMAggNet is able to achieve similar correctness to Ensemble for $\epsilon \in \{0.30, 1.00\}$, with a lower number of incorrect classifications at these high correctness values. For $\epsilon = 3.00$, RMAggNet shows notably higher correctness with significantly lower incorrectness as it is able to leverage the reject option at high $EC$ values. If we focus on minimising incorrectness instead, we can see that Ensemble is able to perform slightly better across all $\epsilon$, however, RMAggNet is able to achieve higher correctness which increases with larger $\epsilon$.}{}

The performance on adversarial datasets generated with open-box attacks is shown in tables \ref{tab:emnist_direct_pgdlinf} and \ref{tab:emnist_direct_pgdl2}. For both of these experiments we aim to minimise incorrectness, either by correctly classifying or rejecting the data. Correctly classifying the data is preferred since it reduces the reliance on downstream rejection handling.

Table \ref{tab:emnist_direct_pgdlinf} shows the results of the PGD $L_\infty$ attack at varying perturbation budgets ($\epsilon$). The CCAT results demonstrate strong performance with 0\% incorrectness for $\tau > 0$ for all $\epsilon$. At $\tau=0$ we effectively disable the confidence threshold of CCAT and see 100\% incorrectness as it is trained to return a uniform distribution for adversarial inputs. It is worth noting that the 0\% incorrectness of CCAT is achieved through the rejection of all of the inputs, even at lower $\epsilon$ where both Ensemble and RMAggNet show that correct classifications can be recovered. This points towards a disadvantage of the conservative nature of CCAT. In situations where we want the option to reject, but can tolerate some incorrectness, CCAT often becomes ineffective for classification.
Comparing Ensemble and RMAggNet, RMAggNet can achieve significantly lower incorrectness over all $\epsilon$, translating the incorrectness into correct or rejected classifications depending on the amount of EC. This leads to RMAggNet being able to achieve higher correctness than both Ensemble and CCAT.

The results of the PGD $L_2$ attacks are in table \ref{tab:emnist_direct_pgdl2}. The results are similar to those in table \ref{tab:emnist_direct_pgdlinf}, with CCAT reducing incorrectness to 0\% through rejection alone. RMAggNet outperforms Ensemble in both maximum correctness and minimum incorrectness for $\epsilon=0.30$ and $\epsilon=1.0$. However, Ensemble can achieve higher correctness for $\epsilon=3.0$ at the cost of incorrectness, which remains significantly higher than RMAggNet's. 

From these results, we can conclude that RMAggNet provides more flexibility where small amounts of incorrectness is tolerable. The error correction process allows us to make trade-offs between maximising correctness and minimising incorrectness, with RMAggNet outperforming Ensemble in both of these metrics. RMAggNet comes close to CCAT in minimising incorrectness with the added advantage that, for small $\epsilon$, we can recover and correctly classify many of the inputs, reducing pressure on downstream rejection handling.

\longpaper{A selection of adversarial images generated using PGD $L_\infty$ and  $L_2$ on the EMNIST dataset can be found in figures \ref{fig:EMNIST_PGDLinf_adv_images} and \ref{fig:EMNIST_PGDL2_images}.}{}

\longpaper{\begin{figure}
    \begin{subfigure}{0.3\textwidth}
    
    \tikzset{every picture/.style={line width=0.75pt}}
        \begin{tikzpicture}[x=0.75pt,y=0.75pt,yscale=-1,xscale=1]
        
        \draw (-45,0) node [anchor=north west][inner sep=0.75pt]    {Surrogate Model};
        \draw (0,30) node  {\includegraphics[width=1.75\linewidth]{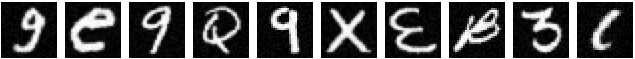}};
        \draw (0,60) node  {\includegraphics[width=1.75\linewidth]{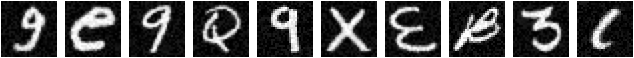}};
        \draw (0,90) node  {\includegraphics[width=1.75\linewidth]{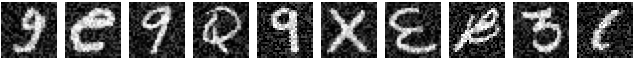}};
        
        \draw (255,0) node [anchor=north west][inner sep=0.75pt]    {CCAT Model};
        \draw (300,30) node  {\includegraphics[width=1.75\linewidth]{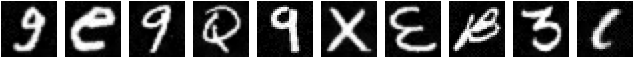}};
        \draw (300,60) node  {\includegraphics[width=1.75\linewidth]{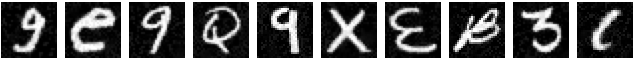}};
        \draw (300,90) node  {\includegraphics[width=1.75\linewidth]{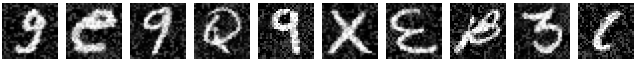}};

        \draw (-45,120) node [anchor=north west][inner sep=0.75pt]    {Ensemble Model};
        \draw (0,150) node  {\includegraphics[width=1.75\linewidth]{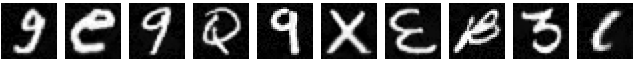}};
        \draw (0,180) node  {\includegraphics[width=1.75\linewidth]{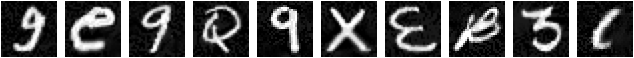}};
        \draw (0,210) node  {\includegraphics[width=1.75\linewidth]{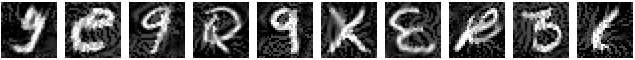}};

        \draw (255,120) node [anchor=north west][inner sep=0.75pt]    {RMAggNet Model};
        \draw (300,150) node  {\includegraphics[width=1.75\linewidth]{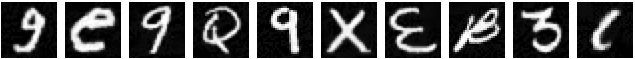}};
        \draw (300,180) node  {\includegraphics[width=1.75\linewidth]{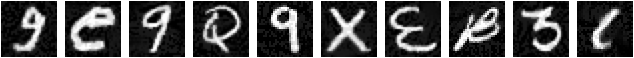}};
        \draw (300,210) node  {\includegraphics[width=1.75\linewidth]{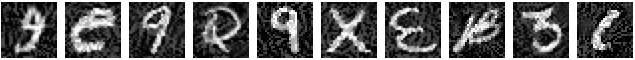}};
        
        \end{tikzpicture}

    \end{subfigure}
    \caption{Adversarial images generated using the PGD $L_\infty$ attack on the EMNIST dataset across all models. Each architecture shows three rows of images, where each row shows the adversarial images generated at a different $\epsilon \in \{0.05, 0.1, 0.3\}$.}
    \label{fig:EMNIST_PGDLinf_adv_images}
\end{figure}
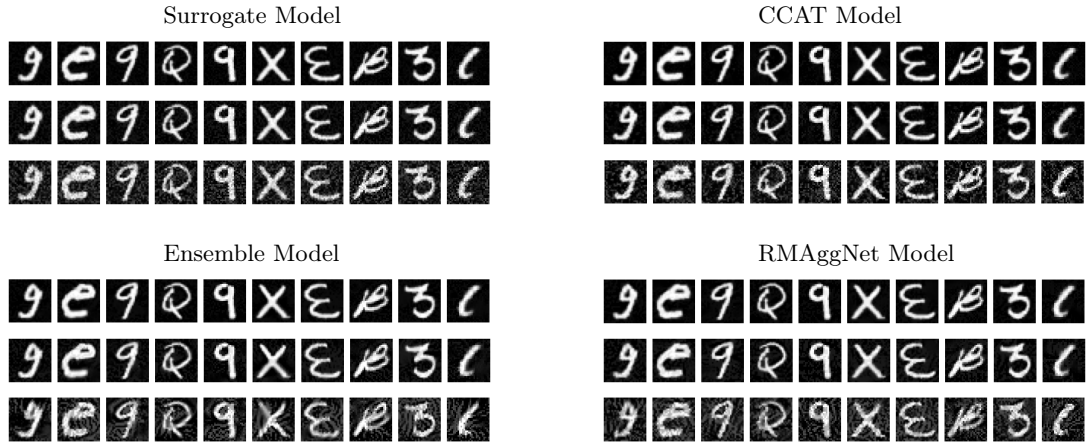

}{}
\longpaper{\begin{table}[htbp]
    \caption{Results for the transfer attacks using PGD $L_\infty$. Table \ref{tab:emnist_surrogate_pgdlinf_res} shows the accuracy of the surrogate model on the PGD $L_\infty$ adversarial datasets. Table \ref{tab:emnist_transfer_pgdlinf} shows the results of the adversarial datasets on the CCAT, Ensemble and RMAggNet models.}
    \centering
    \begin{subtable}[h]{1\textwidth}
        \caption{Accuracy of the surrogate EMNIST classifier on the adversarial datasets generated using PGD $L_\infty$ with different perturbation budgets ($\epsilon$).}
        \label{tab:emnist_surrogate_pgdlinf_res}
        \begin{center}
            \begin{tabular}{|c|c|}
                \hline
                 $\epsilon$ & \textbf{Accuracy (\%)} \\
                 \hline
                 0.00 & 87.93 \\
                 0.05 & 2.20  \\
                 0.10 & 0.00  \\
                 0.30 & 0.00  \\ 
                 \hline
            \end{tabular}%
        \end{center}
        
    \end{subtable}
    
    \centering
    \begin{subtable}[h]{\columnwidth}
    \caption{Percentage of correct, rejected and incorrect classifications of the models using transfer attacks on a surrogate EMNIST classifier using the PGD $L_\infty$ attack. Lower incorrectness is better.}
    \label{tab:emnist_transfer_pgdlinf}
    \resizebox{1\columnwidth}{!}{%
    \begin{tabular}{|*{13}{c|}}
        \hline
        \multicolumn{13}{|c|}{\textbf{PGD($L_\infty$)}}\\
        \hline
        \multicolumn{13}{|l|}{\textbf{CCAT}}\\
        \hline
          $\tau$ & $\epsilon$ & \textbf{Correct} & \textbf{Rejected} & \textbf{Incorrect} & $\epsilon$ & \textbf{Correct} & \textbf{Rejected} & \textbf{Incorrect} & $\epsilon$ & \textbf{Correct} & \textbf{Rejected} & \textbf{Incorrect} \\
         \hline
          0.00 & 0.05 & 3.40 & 0.00   & \textbf{96.60} & 0.10 & 2.90 & 0.00   & \textbf{97.10} & 0.30 & 2.30 & 0.00   & \textbf{97.70}  \\
          0.30 &      & 0.00 & 100.00 & \textbf{0.00}  &      & 0.00 & 100.00 & \textbf{0.00}  &      & 0.00 & 100.00 & \textbf{0.00}   \\
          0.70 &      & 0.00 & 100.00 & \textbf{0.00}  &      & 0.00 & 100.00 & \textbf{0.00}  &      & 0.00 & 100.00 & \textbf{0.00}   \\
          1.00 &      & 0.00 & 100.00 & \textbf{0.00}  &      & 0.00 & 100.00 & \textbf{0.00}  &      & 0.00 & 100.00 & \textbf{0.00}   \\
          \hline
          \multicolumn{13}{|l|}{\textbf{Ensemble}}\\
          \hline
          $\sigma$ & $\epsilon$ & \textbf{Correct} & \textbf{Rejected} & \textbf{Incorrect} & $\epsilon$ & \textbf{Correct} & \textbf{Rejected} & \textbf{Incorrect} & $\epsilon$ & \textbf{Correct} & \textbf{Rejected} & \textbf{Incorrect} \\
          \hline
          0.00 & 0.05 & 88.60 & 0.00  & \textbf{11.40} & 0.10 & 85.40 & 0.00  & \textbf{14.60} & 0.30 & 30.60 & 0.00  & \textbf{69.40}  \\ 
          0.30 &      & 88.60 & 0.00  & \textbf{11.40} &      & 85.40 & 0.20  & \textbf{14.40} &      & 29.90 & 6.30  & \textbf{63.80}  \\ 
          0.70 &      & 84.30 & 9.30  & \textbf{6.40}  &      & 80.30 & 11.10 & \textbf{8.60}  &      & 18.00 & 58.00 & \textbf{24.00}  \\
          1.00 &      & 63.30 & 35.10 & \textbf{1.60}  &      & 50.20 & 48.00 & \textbf{1.80}  &      & 3.80  & 93.60 & \textbf{2.60}   \\
          \hline
          \multicolumn{13}{|l|}{\textbf{RMAggNet}}\\
          \hline
          EC & $\epsilon$ & \textbf{Correct} & \textbf{Rejected} & \textbf{Incorrect} & $\epsilon$ & \textbf{Correct} & \textbf{Rejected} & \textbf{Incorrect} & $\epsilon$ & \textbf{Correct} & \textbf{Rejected} & \textbf{Incorrect} \\
          \hline
          0 & 0.05 & 66.40 & 31.60 & \textbf{2.00} & 0.10 & 60.00 & 37.90 & \textbf{2.10} & 0.30 & 7.30  & 91.20 & \textbf{1.50}   \\ 
          1 &      & 74.50 & 22.50 & \textbf{3.00} &      & 70.10 & 26.70 & \textbf{3.20} &      & 14.10 & 81.80 & \textbf{4.10}   \\ 
          2 &      & 79.20 & 17.20 & \textbf{3.60} &      & 76.00 & 19.60 & \textbf{4.40} &      & 18.80 & 74.20 & \textbf{7.00}   \\ 
          3 &      & 82.70 & 12.80 & \textbf{4.50} &      & 79.60 & 15.50 & \textbf{4.90} &      & 24.70 & 66.10 & \textbf{9.20}   \\ 
          4 &      & 84.40 & 10.10 & \textbf{5.50} &      & 81.50 & 12.50 & \textbf{6.00} &      & 28.60 & 59.80 & \textbf{11.60}  \\ 
          5 &      & 85.70 & 7.90  & \textbf{6.40} &      & 83.00 & 9.80  & \textbf{7.20} &      & 33.40 & 50.80 & \textbf{15.80}  \\ 
          6 &      & 87.60 & 5.20  & \textbf{7.20} &      & 85.00 & 7.00  & \textbf{8.00} &      & 39.00 & 40.00 & \textbf{21.00}  \\ 
          7 &      & 89.40 & 2.60  & \textbf{8.00} &      & 87.50 & 2.80  & \textbf{9.70} &      & 44.10 & 28.50 & \textbf{27.40}  \\ 
          \hline
    \end{tabular}%
    }
    
    \end{subtable}
    
\end{table}
}{}
\begin{table}[!htbp]
  \caption{Results of PGD $L_\infty$ adversaries generated using open-box attacks on EMNIST images with percentages of correct, rejected and incorrect classifications. Lower incorrectness is better.}
    \label{tab:emnist_direct_pgdlinf}
    \centering
    \resizebox{1\columnwidth}{!}{%
    \begin{tabular}{|*{13}{c|}}
        \hline
        \multicolumn{13}{|c|}{\textbf{PGD($L_\infty$)}}\\
        \hline
        \multicolumn{13}{|l|}{\textbf{CCAT}}\\
        \hline
          $\tau$ & $\epsilon$ & \textbf{Correct} & \textbf{Rejected} & \textbf{Incorrect} & $\epsilon$ & \textbf{Correct} & \textbf{Rejected} & \textbf{Incorrect} & $\epsilon$ & \textbf{Correct} & \textbf{Rejected} & \textbf{Incorrect} \\
         \hline
          0.00 & 0.05 & 0.00 & 0.00   & \textbf{100.00} & 0.10 & 0.00 & 0.00   & \textbf{100.00} & 0.30 & 0.00 & 0.00   & \textbf{100.00}  \\
          0.30 &      & 0.00 & 100.00 & \textbf{0.00}   &      & 0.00 & 100.00 & \textbf{0.00}   &      & 0.00 & 100.00 & \textbf{0.00}  \\
          0.70 &      & 0.00 & 100.00 & \textbf{0.00}   &      & 0.00 & 100.00 & \textbf{0.00}   &      & 0.00 & 100.00 & \textbf{0.00}  \\
          0.90 &      & 0.00 & 100.00 & \textbf{0.00}   &      & 0.00 & 100.00 & \textbf{0.00}   &      & 0.00 & 100.00 & \textbf{0.00}    \\
          \hline
          \multicolumn{13}{|l|}{\textbf{Ensemble}}\\
          \hline
          $\sigma$ & $\epsilon$ & \textbf{Correct} & \textbf{Rejected} & \textbf{Incorrect} & $\epsilon$ & \textbf{Correct} & \textbf{Rejected} & \textbf{Incorrect} & $\epsilon$ & \textbf{Correct} & \textbf{Rejected} & \textbf{Incorrect} \\
          \hline
          0.00 & 0.05 & 71.70 & 0.00  & \textbf{28.30}& 0.10 & 29.40 & 0.00  & \textbf{70.60}& 0.30 & 0.00 & 0.00 & \textbf{100.00} \\ 
          0.30 &      & 71.70 & 0.00  & \textbf{28.30}&      & 29.40 & 0.00  & \textbf{70.60}&      & 0.00 & 0.00 & \textbf{100.00} \\ 
          0.70 &      & 64.10 & 15.20 & \textbf{20.70}&      & 20.80 & 22.60 & \textbf{56.60}&      & 0.00 & 0.00 & \textbf{100.00} \\
          1.00 &      & 31.40 & 60.10 & \textbf{8.50}&      & 3.20  & 73.70 & \textbf{23.10}&      & 0.00 & 4.60 & \textbf{95.40} \\
          \hline
          \multicolumn{13}{|l|}{\textbf{RMAggNet}}\\
          \hline
          EC & $\epsilon$ & \textbf{Correct} & \textbf{Rejected} & \textbf{Incorrect} & $\epsilon$ & \textbf{Correct} & \textbf{Rejected} & \textbf{Incorrect} & $\epsilon$ & \textbf{Correct} & \textbf{Rejected} & \textbf{Incorrect} \\
          \hline
          0 & 0.05 & 28.50 & 67.10 & \textbf{4.40 } & 0.10 & 2.30  & 92.00 & \textbf{5.70 } & 0.30 & 0.00 & 100.00 & \textbf{0.00 }  \\ 
          1 &      & 54.30 & 39.60 & \textbf{6.10 } &      & 16.80 & 72.40 & \textbf{10.80} &      & 0.00 & 98.50  & \textbf{1.50 }  \\ 
          2 &      & 63.20 & 29.10 & \textbf{7.70 } &      & 29.30 & 56.70 & \textbf{14.00} &      & 0.00 & 94.40  & \textbf{5.60 }  \\ 
          3 &      & 68.90 & 22.20 & \textbf{8.90 } &      & 41.40 & 41.50 & \textbf{17.10} &      & 0.10 & 87.90  & \textbf{12.00}  \\
          4 &      & 71.90 & 17.90 & \textbf{10.20} &      & 48.20 & 31.50 & \textbf{20.30} &      & 0.10 & 77.80  & \textbf{22.10}  \\ 
          5 &      & 75.00 & 13.70 & \textbf{11.30} &      & 54.10 & 23.40 & \textbf{22.50} &      & 0.30 & 68.30  & \textbf{31.40}  \\ 
          6 &      & 78.00 & 8.50  & \textbf{13.50} &      & 59.30 & 15.60 & \textbf{25.10} &      & 0.40 & 54.00  & \textbf{45.60}  \\
          7 &      & 80.70 & 3.60  & \textbf{15.70} &      & 62.90 & 8.70  & \textbf{28.40} &      & 0.50 & 39.50  & \textbf{60.00}  \\
          \hline
    \end{tabular}%
    }
    
\end{table}

\longpaper{\begin{figure}
    \begin{subfigure}{0.3\textwidth}
    \centering
    \tikzset{every picture/.style={line width=0.75pt}} 
        \begin{tikzpicture}[x=0.75pt,y=0.75pt,yscale=-1,xscale=1]
        
        \draw (-45,0) node [anchor=north west][inner sep=0.75pt]    {Surrogate Model};
        \draw (0,30) node  {\includegraphics[width=1.75\linewidth]{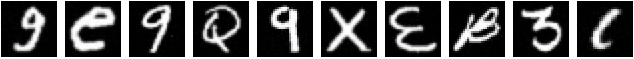}};
        \draw (0,60) node  {\includegraphics[width=1.75\linewidth]{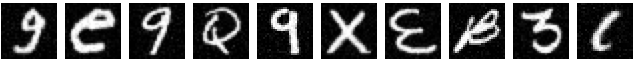}};
        \draw (0,90) node  {\includegraphics[width=1.75\linewidth]{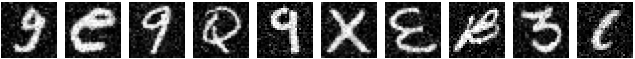}};

        \draw (255,0) node [anchor=north west][inner sep=0.75pt]    {CCAT Model};
        \draw (300,30) node  {\includegraphics[width=1.75\linewidth]{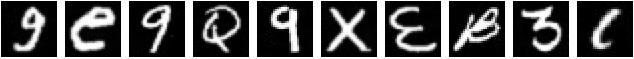}};
        \draw (300,60) node  {\includegraphics[width=1.75\linewidth]{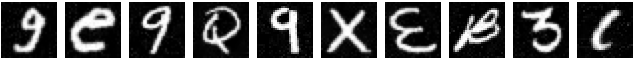}};
        \draw (300,90) node  {\includegraphics[width=1.75\linewidth]{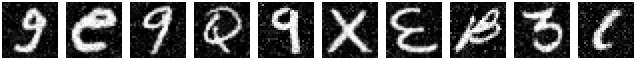}};
        
        \draw (-45,120) node [anchor=north west][inner sep=0.75pt]    {Ensemble Model};
        \draw (0,150) node  {\includegraphics[width=1.75\linewidth]{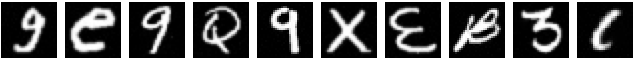}};
        \draw (0,180) node  {\includegraphics[width=1.75\linewidth]{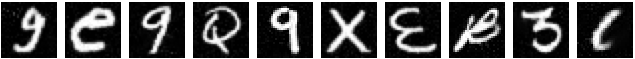}};
        \draw (0,210) node  {\includegraphics[width=1.75\linewidth]{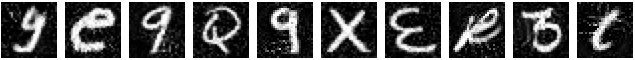}};

        \draw (255,120) node [anchor=north west][inner sep=0.75pt]    {RMAggNet Model};
        \draw (300,150) node  {\includegraphics[width=1.75\linewidth]{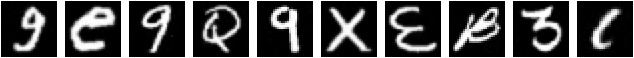}};
        \draw (300,180) node  {\includegraphics[width=1.75\linewidth]{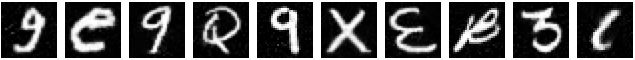}};
        \draw (300,210) node  {\includegraphics[width=1.75\linewidth]{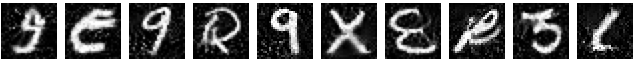}};

        \end{tikzpicture}
    
    \end{subfigure}
    \caption{Adversarial images generated using the PGD $L_2$ attack on the EMNIST dataset across all models. Each architecture shows three rows of images, where each row shows the adversarial images generated at a different $\epsilon \in \{0.30, 1.0, 3.0\}$.}
    \label{fig:EMNIST_PGDL2_images}
\end{figure}

}{}
\longpaper{\begin{table}[htbp]
    \caption{Results for the transfer attacks using PGD $L_2$. Table \ref{tab:emnist_surrogate_res_pgdl2} shows the accuracy of the surrogate model on the PGD $L_2$ adversarial datasets. Table \ref{tab:emnist_transfer_pgdl2} shows the results of the adversarial datasets on the CCAT, Ensemble and RMAggNet models.}
    \centering
    \begin{subtable}[h]{1\textwidth}
        \caption{Accuracy of the surrogate EMNIST classifier on the adversarial datasets generated using PGD $L_2$ with different perturbation budgets ($\epsilon$).}
        \label{tab:emnist_surrogate_res_pgdl2}
        \begin{center}
            \begin{tabular}{|c|c|}
                \hline
                 $\epsilon$ & \textbf{Accuracy (\%)} \\
                 \hline
                 0.00 & 87.93 \\
                 0.30 & 55.10 \\
                 1.00 & 1.20  \\
                 3.00 & 0.00  \\
                 \hline
            \end{tabular}%
        \end{center}
        
    \end{subtable}
    
    \begin{subtable}[h]{\textwidth}
    \caption{Percentage of correct, rejected and incorrect classifications of the models using transfer attacks on a surrogate EMNIST classifier using the PGD $L_2$ attack. Lower incorrectness is better.}
    \label{tab:emnist_transfer_pgdl2}
    \resizebox{1\columnwidth}{!}{%
    \begin{tabular}{|*{13}{c|}}
        \hline
        \multicolumn{13}{|c|}{\textbf{PGD($L_2$)}}\\
        \hline
        \multicolumn{13}{|l|}{\textbf{CCAT}}\\
        \hline
          $\tau$ & $\epsilon$ & \textbf{Correct} & \textbf{Rejected} & \textbf{Incorrect} & $\epsilon$ & \textbf{Correct} & \textbf{Rejected} & \textbf{Incorrect} & $\epsilon$ & \textbf{Correct} & \textbf{Rejected} & \textbf{Incorrect} \\
         \hline
          0.00 & 0.30 & 30.20 & 0.00   & \textbf{69.80} & 1.00 & 3.00 & 0.00   & \textbf{97.00} & 3.00 & 2.70 & 0.00   & \textbf{97.30} \\
          0.30 &      & 0.00  & 100.00 & \textbf{0.00}  &      & 0.00 & 100.00 & \textbf{0.00}  &      & 0.00 & 100.00 & \textbf{0.00}  \\
          0.70 &      & 0.00  & 100.00 & \textbf{0.00}  &      & 0.00 & 100.00 & \textbf{0.00}  &      & 0.00 & 100.00 & \textbf{0.00}  \\
          1.00 &      & 0.00  & 100.00 & \textbf{0.00}  &      & 0.00 & 100.00 & \textbf{0.00}  &      & 0.00 & 100.00 & \textbf{0.00}  \\
          \hline
          \multicolumn{13}{|l|}{\textbf{Ensemble}}\\
          \hline
          $\sigma$ & $\epsilon$ & \textbf{Correct} & \textbf{Rejected} & \textbf{Incorrect} & $\epsilon$ & \textbf{Correct} & \textbf{Rejected} & \textbf{Incorrect} & $\epsilon$ & \textbf{Correct} & \textbf{Rejected} & \textbf{Incorrect} \\
          \hline
          0.00 & 0.30 & 90.00 & 0.00  & \textbf{10.00} & 1.00 & 87.50 & 0.00  & \textbf{12.50} & 3.00 & 64.40 & 0.00  & \textbf{35.60} \\
          0.30 &      & 90.00 & 0.00  & \textbf{10.00} &      & 87.50 & 0.10  & \textbf{12.40} &      & 64.40 & 1.50  & \textbf{34.10} \\
          0.70 &      & 85.80 & 8.60  & \textbf{5.60}  &      & 83.00 & 9.80  & \textbf{7.20}  &      & 45.90 & 39.30 & \textbf{14.80} \\
          1.00 &      & 66.60 & 32.20 & \textbf{1.20}  &      & 59.50 & 38.70 & \textbf{1.80}  &      & 17.30 & 81.40 & \textbf{1.30}  \\
          \hline
          \multicolumn{13}{|l|}{\textbf{RMAggNet}}\\
          \hline
          EC & $\epsilon$ & \textbf{Correct} & \textbf{Rejected} & \textbf{Incorrect} & $\epsilon$ & \textbf{Correct} & \textbf{Rejected} & \textbf{Incorrect} & $\epsilon$ & \textbf{Correct} & \textbf{Rejected} & \textbf{Incorrect} \\
          \hline
          0 & 0.30 & 68.90 & 29.20 & \textbf{1.90} & 1.00 & 65.40 & 32.20 & \textbf{2.40} & 3.00 & 26.10 & 72.60 & \textbf{1.30}  \\
          1 &      & 76.30 & 20.70 & \textbf{3.00} &      & 73.10 & 24.00 & \textbf{2.90} &      & 36.00 & 61.20 & \textbf{2.80}  \\
          2 &      & 81.60 & 15.20 & \textbf{3.20} &      & 77.70 & 18.60 & \textbf{3.70} &      & 46.40 & 49.10 & \textbf{4.50}  \\
          3 &      & 84.00 & 11.60 & \textbf{4.40} &      & 80.90 & 15.00 & \textbf{4.10} &      & 53.00 & 39.50 & \textbf{7.50}  \\
          4 &      & 85.70 & 9.40  & \textbf{4.90} &      & 82.90 & 11.40 & \textbf{5.70} &      & 59.30 & 31.50 & \textbf{9.20}  \\
          5 &      & 87.00 & 7.10  & \textbf{5.90} &      & 84.60 & 9.10  & \textbf{6.30} &      & 65.00 & 24.00 & \textbf{11.00} \\
          6 &      & 88.30 & 5.40  & \textbf{6.30} &      & 86.70 & 5.70  & \textbf{7.60} &      & 69.30 & 17.40 & \textbf{13.30} \\
          7 &      & 89.60 & 2.70  & \textbf{7.70} &      & 88.30 & 2.60  & \textbf{9.10} &      & 73.90 & 9.40  & \textbf{16.70} \\
          \hline
    \end{tabular}%
    }
    
    \end{subtable}
\end{table}
}{}
\begin{table}[!htbp]
    \caption{Results of PGD $L_2$ adversaries generated using open-box attacks on EMNIST images with percentages of correct, rejected and incorrect classifications. Lower incorrectness is better.}
    \label{tab:emnist_direct_pgdl2}
    \centering
    \resizebox{1\columnwidth}{!}{%
    \begin{tabular}{|*{13}{c|}}
        \hline
        \multicolumn{13}{|c|}{\textbf{PGD($L_2$)}}\\
        \hline
        \multicolumn{13}{|l|}{\textbf{CCAT}}\\
        \hline
          $\tau$ & $\epsilon$ & \textbf{Correct} & \textbf{Rejected} & \textbf{Incorrect} & $\epsilon$ & \textbf{Correct} & \textbf{Rejected} & \textbf{Incorrect} & $\epsilon$ & \textbf{Correct} & \textbf{Rejected} & \textbf{Incorrect} \\
         \hline
          0.00 & 0.30 & 0.20 & 0.00   & \textbf{99.80 }&  1.0 & 0.00 & 0.00   & \textbf{100.00} & 3.0 & 0.00 & 0.00   & \textbf{100.00}  \\
          0.30 &      & 0.00 & 100.00 & \textbf{0.00  }&      & 0.00 & 100.00 & \textbf{0.00  } &     & 0.00 & 100.00 & \textbf{0.00  }  \\
          0.70 &      & 0.00 & 100.00 & \textbf{0.00  }&      & 0.00 & 100.00 & \textbf{0.00  } &     & 0.00 & 100.00 & \textbf{0.00  }  \\
          0.90 &      & 0.00 & 100.00 & \textbf{0.00}  &      & 0.00 & 100.00 & \textbf{0.00}   &     & 0.00 & 100.00 & \textbf{0.00}    \\
          \hline
          \multicolumn{13}{|l|}{\textbf{Ensemble}}\\
          \hline
          $\sigma$ & $\epsilon$ & \textbf{Correct} & \textbf{Rejected} & \textbf{Incorrect} & $\epsilon$ & \textbf{Correct} & \textbf{Rejected} & \textbf{Incorrect} & $\epsilon$ & \textbf{Correct} & \textbf{Rejected} & \textbf{Incorrect} \\
          \hline
          0.00 & 0.30 & 84.80 & 0.00  & \textbf{15.20} &  1.0 & 62.40 & 0.00  & \textbf{37.60} & 3.0 & 19.60 & 0.00 & \textbf{80.40}  \\
          0.30 &      & 84.80 & 0.00  & \textbf{15.20} &      & 62.40 & 0.00  & \textbf{37.60} &     & 19.60 & 0.00 & \textbf{80.40}  \\
          0.70 &      & 79.50 & 9.10  & \textbf{11.40} &      & 57.10 & 13.00 & \textbf{29.90} &     & 19.50 & 0.20 & \textbf{80.30}  \\
          1.00 &      & 60.60 & 35.90 & \textbf{3.50 } &      & 47.40 & 39.80 & \textbf{12.80} &     & 18.50 & 9.80 & \textbf{71.70}  \\
          \hline
          \multicolumn{13}{|l|}{\textbf{RMAggNet}}\\
          \hline
          EC & $\epsilon$ & \textbf{Correct} & \textbf{Rejected} & \textbf{Incorrect} & $\epsilon$ & \textbf{Correct} & \textbf{Rejected} & \textbf{Incorrect} & $\epsilon$ & \textbf{Correct} & \textbf{Rejected} & \textbf{Incorrect}\\
          \hline
          0 & 0.30 & 56.00 & 41.30 & \textbf{2.70 } & 1.0 & 15.10 & 78.60 & \textbf{6.30 } & 3.0 & 0.00 & 94.60 & \textbf{5.40 }  \\
          1 &      & 68.00 & 28.20 & \textbf{3.80 } &     & 39.10 & 51.50 & \textbf{9.40 } &     & 0.70 & 86.10 & \textbf{13.20}  \\
          2 &      & 74.10 & 20.90 & \textbf{5.00 } &     & 50.30 & 38.10 & \textbf{11.60} &     & 1.00 & 75.60 & \textbf{23.40}  \\
          3 &      & 77.70 & 16.40 & \textbf{5.90 } &     & 57.40 & 29.30 & \textbf{13.30} &     & 2.00 & 65.00 & \textbf{33.00}  \\
          4 &      & 80.70 & 12.40 & \textbf{6.90 } &     & 61.70 & 23.40 & \textbf{14.90} &     & 3.10 & 55.60 & \textbf{41.30}  \\
          5 &      & 82.70 & 9.10  & \textbf{8.20 } &     & 65.20 & 17.40 & \textbf{17.40} &     & 5.00 & 45.60 & \textbf{49.40}  \\
          6 &      & 84.40 & 6.20  & \textbf{9.40 } &     & 67.60 & 12.20 & \textbf{20.20} &     & 6.40 & 37.60 & \textbf{56.00}  \\
          7 &      & 86.00 & 3.10  & \textbf{10.90} &     & 70.40 & 6.60  & \textbf{23.00} &     & 9.20 & 25.70 & \textbf{65.10}  \\
          \hline
    \end{tabular}%
    }
    
\end{table}

\subsection{CIFAR-10 Dataset} \label{sec:res_cifar}

Results for the CIFAR-10 dataset use RMAggNet with $m=4$, $r=1$ which gives us 16 networks with 3 bits of error correction. We use 16 networks in the Ensemble method for parity. All networks use an architecture outlined in \longpaper{table \ref{tab:cifar_arch}.}{the extended paper (table 17).}

\longpaper{\begin{table}[htbp]
    \caption{Model architecture for classifying CIFAR-10 data. The $c$ parameter is for the varying output sizes depending on the system i.e. $c=1$ for RMAggNet, $c=10$ for Ensemble and CCAT.}
    \label{tab:cifar_arch}
    \centering
    \begin{tabular}{r|l}
        \textbf{Layer} & \textbf{Parameters} \\
        \hline
        Conv2D & channels = 32, kernel size = $5 \times 5$, padding = 2 \\
        BatchNorm2D &  \\
        ReLU &  \\
        Conv2D & channels = 32, kernel size = $5 \times 5$, padding = 2 \\
        BatchNorm2D &  \\
        ReLU &  \\
        
        MaxPool2D & pool size = $2\times 2$ \\
        Dropout & $p=0.4$ \\

        Conv2D & channels = 64, kernel size = $5 \times 5$, padding = 2 \\
        BatchNorm2D &  \\
        ReLU &  \\
        Conv2D & channels = 64, kernel size = $5 \times 5$, padding = 2 \\
        BatchNorm2D &  \\
        ReLU &  \\

        MaxPool2D & pool size = $2\times 2$ \\
        Dropout & $p=0.5$ \\

        Conv2D & channels = 128, kernel size = $5 \times 5$, padding = 2 \\
        BatchNorm2D &  \\
        ReLU &  \\
        Conv2D & channels = 128, kernel size = $5 \times 5$, padding = 2 \\
        BatchNorm2D &  \\
        ReLU &  \\

        MaxPool2D & pool size = $2\times 2$ \\
        Dropout & $p=0.6$ \\

        Flatten & \\

        Linear & in = 2048, out = 512 \\
        ReLU & \\
        Dropout & $p=0.7$ \\

        Linear & in = 512, out = 256 \\
        ReLU & \\
        Dropout & $p=0.8$ \\

        Linear & in = 256, out = $c$ \\
        
    \end{tabular}
    
\end{table}}{}

Table \ref{tab:cifar10_res} shows the results on the clean CIFAR-10 dataset where we aim to maximise correctness. Ensemble reports the highest correctness, followed by RMAggNet, then CCAT. All models report correctness within 3\% of one another, therefore, we can conclude that all are equally capable in terms of classification ability.

\begin{table}[!htbp]
    \caption{Results for the clean CIFAR-10 dataset showing the percentage of classifications that are correct, rejected and incorrect. Bold text indicates the metric of interest. Higher correctness is better.}
    \label{tab:cifar10_res}
    \begin{subtable}[h]{0.32938\columnwidth}
    \resizebox{1\columnwidth}{!}{
        \begin{tabular}{|*{4}{c|}}
            \hline
            \multicolumn{4}{|l|}{\textbf{CCAT}}\\
            \hline
            \textbf{$\tau$} & \textbf{Correct} & \textbf{Rejected} & \textbf{Incorrect} \\
            \hline
            0    &  \textbf{76.41} &  0.00 & 23.59 \\
            0.10 &  \textbf{76.41} &  0.00 & 23.59 \\
            0.20 &  \textbf{76.21} &  0.68 & 23.11 \\
            0.30 &  \textbf{75.10} &  3.76 & 21.14 \\
            0.40 &  \textbf{72.62} &  9.38 & 18.00 \\
            0.50 &  \textbf{68.59} & 17.41 & 14.00 \\
            0.60 &  \textbf{64.04} & 25.55 & 10.41 \\
            0.70 &  \textbf{58.81} & 33.72 &  7.47 \\
            0.80 &  \textbf{52.81} & 42.39 &  4.80 \\
            0.90 &  \textbf{44.64} & 52.54 &  2.82 \\
            1.0  &  \textbf{ 1.09} & 98.91 &  0.00 \\
            \hline
        \end{tabular}
        }
        \label{tab:cifar10_ccat}
    \end{subtable}
    \begin{subtable}[h]{0.32938\columnwidth}
        \resizebox{1\columnwidth}{!}{
        \begin{tabular}{|*{4}{c|}}
            \hline
            \multicolumn{4}{|l|}{\textbf{Ensemble}}\\
            \hline
            \textbf{$\sigma$} & \textbf{Correct} & \textbf{Rejected} & \textbf{Incorrect} \\
            \hline
            0    & \textbf{79.34} & 0.00 & 20.66 \\
            0.10 & \textbf{79.54} & 0.00 & 20.46 \\
            0.20 & \textbf{79.35} & 0.00 & 20.65 \\
            0.30 & \textbf{79.53} & 0.02 & 20.45 \\
            0.40 & \textbf{79.24} & 1.05 & 19.71 \\
            0.50 & \textbf{77.14} & 6.73 & 16.13 \\
            0.60 & \textbf{75.31} & 11.23 & 13.46 \\
            0.70 & \textbf{68.79} & 22.62 & 8.59 \\
            0.80 & \textbf{65.38} & 28.00 & 6.62 \\
            0.90 & \textbf{56.57} & 40.23 & 3.20 \\
            1.0  & \textbf{39.21} & 59.83 & 0.96 \\
            \hline
        \end{tabular}
        }
        \label{tab:cifar10_ensemble}
    \end{subtable}    
    \begin{subtable}[h]{0.32938\columnwidth}
        \resizebox{1\columnwidth}{!}{
        \begin{tabular}{|*{4}{c|}}
            \hline
            \multicolumn{4}{|l|}{\textbf{RMAggNet} $[16, 5, 8]_2$}\\
            \hline
            \textbf{EC} & \textbf{Correct} & \textbf{Rejected} & \textbf{Incorrect} \\
            \hline
            0 & \textbf{42.46} & 59.96 & 0.58 \\
            1 & \textbf{57.90} & 40.09 & 2.01 \\
            2 & \textbf{68.32} & 27.23 & 4.45 \\
            3 & \textbf{77.11} & 12.76 & 10.13 \\
            \hline
        \end{tabular}
        }
        \label{tab:cifar10_rmaggnet}
    \end{subtable}
\end{table}

\longpaper{The $L_\infty$ transfer attacks on the CIFAR-10 models can be seen in table \ref{tab:cifar_transfer_pgdlinf}, where we aim to reduce incorrectness. The results show a strong adversarial attack due to low correctness from Ensemble and RMAggNet, even at low $\epsilon$. The CCAT model is able to reliably reject all adversarial inputs leading to 0\% incorrectness for all $\epsilon$ at all non-zero confidence thresholds. The Ensemble method classifies significantly more inputs correctly compared to RMAggNet for all $\epsilon$ we tested, however, these are still a small number of correct classifications. This result shows that CCAT is an effective method for avoiding adversarial results when strong attacks are used.}{}

\longpaper{The results of the transfer attacks using the PGD $L_2$ adversarial method are in table \ref{tab:cifar_transfer_pgdl2}. The CCAT method shows some attempt to classify the adversarial images at low $\epsilon$, and it is able to achieve low incorrectness at high $\tau$. However, the lowest incorrectness CCAT reaches at $\epsilon = 0.30$ is 0.6 (apart from $\tau=1$ which rejects all inputs). Comparing this to the closest equivalent from RMAggNet ($EC=0$) we see that RMAggNet is able to have a correctness score which is approximately 20\% higher, while having similar incorrectness, indicating that CCAT is relying on rejection to reduce the incorrectness score whereas RMAggNet attempts to correct the input to the true class. At the lower $\epsilon$ values, Ensemble and RMAggNet are able to achieve similar correctness and incorrectness values, however, Ensemble is able to outperform RMAggNet as $\epsilon$ increases.}{}

The performance on the CIFAR-10 datasets using open-box attacks is shown in tables \ref{tab:cifar_direct_pgdlinf} and \ref{tab:cifar_direct_pgdl2}. We, again, aim to minimise incorrectness.

The PGD $L_\infty$ results in table \ref{tab:cifar_direct_pgdlinf} show low correctness for both Ensemble and RMAggNet for all $\epsilon$ which indicates that this is a strong adversarial attack, which is reduced to unrecognisable images at $\epsilon=0.3$. With this in mind, it is better to compare the methods focusing on incorrectness and rejection performance. Ensemble struggles to reject inputs, leading to nearly 100\% incorrectness across all $\epsilon$ which indicates that the adversarial examples are able to fool the multiple networks that form this method. RMAggNet shows slightly better performance, with much higher rejection for $EC=0$. CCAT can achieve the lowest incorrectness scores, which are significantly lower for $\epsilon=0.75$ and $\epsilon=2.5$. This indicates that when we expect strong adversaries with little chance of recovery, CCAT is the best-performing model.

Table \ref{tab:cifar_direct_pgdl2} shows the results for PGD $L_2$ on CIFAR-10. Interestingly, RMAggNet can outperform both Ensemble and CCAT at $\epsilon = 0.30$ with a lower incorrectness and higher correctness than both methods. For $\epsilon = 0.75$ and $\epsilon=2.5$ CCAT can report the lowest incorrectness by a significant margin. Over all $\epsilon$ values RMAggNet reports lower incorrectness than Ensemble, achieving higher correctness at $\epsilon = \{0.30, 0.75\}$.

These results show the effect that strong adversaries have on the classification ability of these models. In this circumstance, CCAT is the better model, rejecting most adversaries, while Ensemble struggles to reject the inputs, and RMAggNet has varying performance when attempting to correct the images. However, this is a worst-case scenario.

\longpaper{A sample of the adversarial images produced from these attacks can be seen in figures \ref{fig:CIFAR_PGDLinf_adv_images} and \ref{fig:CIFAR_PGDL2_images}.}{}

\longpaper{\begin{figure}
    \begin{subfigure}{0.3\textwidth}
    
    \tikzset{every picture/.style={line width=0.75pt}}
        \begin{tikzpicture}[x=0.75pt,y=0.75pt,yscale=-1,xscale=1]
        
        \draw (-45,0) node [anchor=north west][inner sep=0.75pt]    {Surrogate Model};
        \draw (0,30) node  {\includegraphics[width=1.75\linewidth]{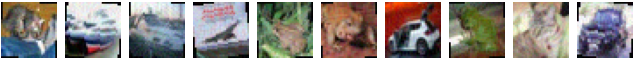}};
        \draw (0,60) node  {\includegraphics[width=1.75\linewidth]{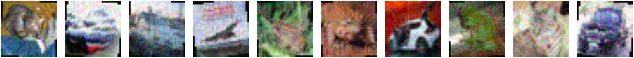}};
        \draw (0,90) node  {\includegraphics[width=1.75\linewidth]{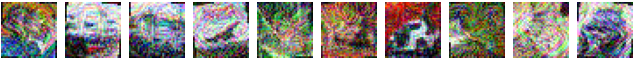}};
        
        \draw (255,0) node [anchor=north west][inner sep=0.75pt]    {CCAT Model};
        \draw (300,30) node  {\includegraphics[width=1.75\linewidth]{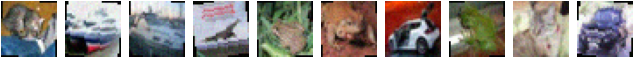}};
        \draw (300,60) node  {\includegraphics[width=1.75\linewidth]{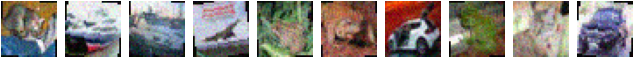}};
        \draw (300,90) node  {\includegraphics[width=1.75\linewidth]{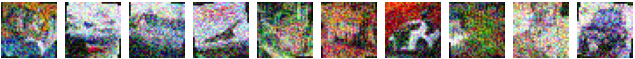}};

        \draw (-45,120) node [anchor=north west][inner sep=0.75pt]    {Ensemble Model};
        \draw (0,150) node  {\includegraphics[width=1.75\linewidth]{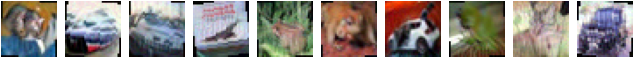}};
        \draw (0,180) node  {\includegraphics[width=1.75\linewidth]{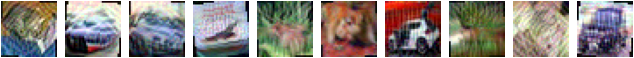}};
        \draw (0,210) node  {\includegraphics[width=1.75\linewidth]{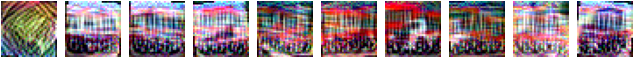}};

        \draw (255,120) node [anchor=north west][inner sep=0.75pt]    {RMAggNet Model};
        \draw (300,150) node  {\includegraphics[width=1.75\linewidth]{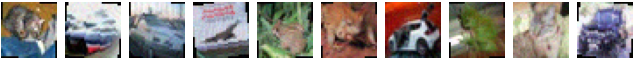}};
        \draw (300,180) node  {\includegraphics[width=1.75\linewidth]{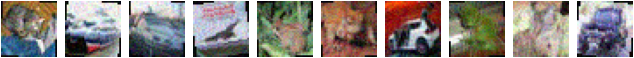}};
        \draw (300,210) node  {\includegraphics[width=1.75\linewidth]{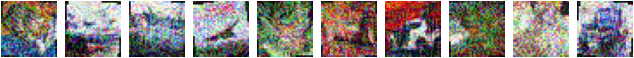}};
        
        \end{tikzpicture}
    
    \end{subfigure}
    \caption{Adversarial images generated using the PGD $L_\infty$ attack on the CIFAR-10 dataset across all models. Each architecture shows three rows of images, where each row shows the adversarial images generated at a different $\epsilon \in \{0.05, 0.1, 0.3\}$.}
    \label{fig:CIFAR_PGDLinf_adv_images}
\end{figure}
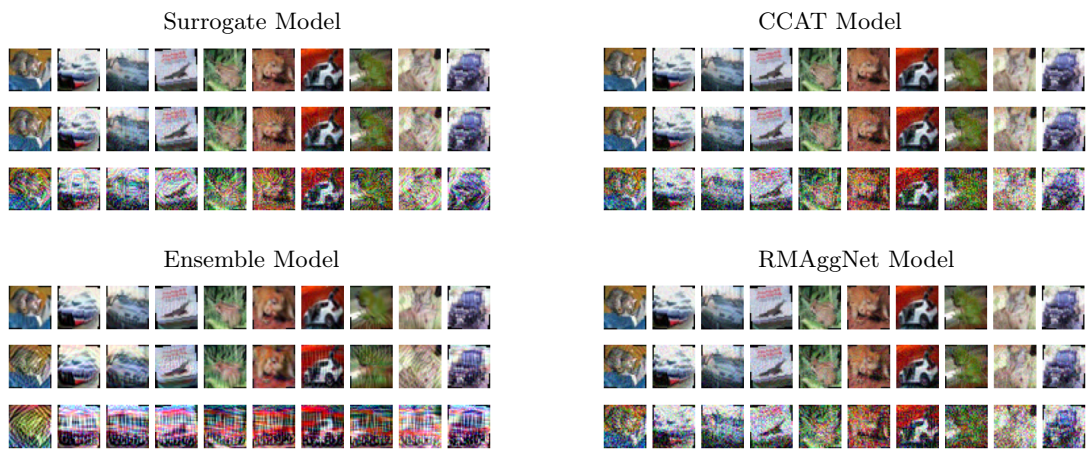

}{}
\longpaper{
\begin{table}[htbp]
    \caption{Results for the transfer attacks using PGD $L_\infty$. Table \ref{tab:cifar_surrogate_pgdlinf_res} shows the accuracy of the surrogate model on the PGD $L_\infty$ adversarial datasets. Table \ref{tab:cifar_transfer_pgdlinf} shows the results of the adversarial datasets on the CCAT, Ensemble and RMAggNet models.}
    \centering
    \begin{subtable}[h]{1\textwidth}
        \caption{Accuracy of the surrogate CIFAR-10 classifier on the adversarial datasets generated using PGD $L_\infty$ with different perturbation budgets ($\epsilon$).}
        \label{tab:cifar_surrogate_pgdlinf_res}
        \begin{center}
            \begin{tabular}{|c|c|}
                \hline
                 $\epsilon$ & \textbf{Accuracy (\%)} \\
                 \hline
                 0.00 & 81.21 \\
                 0.05 & 0.10 \\
                 0.10  & 0.10 \\
                 0.30  & 0.00 \\

                 \hline
            \end{tabular}%
        \end{center}
    \end{subtable}
    
    \begin{subtable}[h]{1\textwidth}
    \caption{Percentage of correct, rejected and incorrect classifications of the models using transfer attacks on a surrogate CIFAR-10 classifier using the PGD $L_\infty$ attack. Lower incorrectness is better.}
    \label{tab:cifar_transfer_pgdlinf}
    \resizebox{1\columnwidth}{!}{%
    \begin{tabular}{|*{13}{c|}}
        \hline
        \multicolumn{13}{|c|}{\textbf{PGD($L_\infty$)}}\\
        \hline
        \multicolumn{13}{|l|}{\textbf{CCAT}}\\
        \hline
          $\tau$ & $\epsilon$ & \textbf{Correct} & \textbf{Rejected} & \textbf{Incorrect} & $\epsilon$ & \textbf{Correct} & \textbf{Rejected} & \textbf{Incorrect} & $\epsilon$ & \textbf{Correct} & \textbf{Rejected} & \textbf{Incorrect} \\
         \hline
          0.00 & 0.05 & 14.80 & 0.00  & \textbf{85.20} & 0.10 & 9.50 & 0.00   & \textbf{90.50} & 0.30 & 11.20  & 0.00   & \textbf{88.80} \\
          0.30 &      & 0.00 & 100.00 & \textbf{0.00}  &      & 0.00 & 100.00 & \textbf{0.00}  &      &  0.00  & 100.00 & \textbf{0.00}  \\
          0.70 &      & 0.00 & 100.00 & \textbf{0.00}  &      & 0.00 & 100.00 & \textbf{0.00}  &      &  0.00  & 100.00 & \textbf{0.00}  \\
          1.00 &      & 0.00 & 100.00 & \textbf{0.00}  &      & 0.00 & 100.00 & \textbf{0.00}  &      &  0.00  & 100.00 & \textbf{0.00}  \\
          \hline
          \multicolumn{13}{|l|}{\textbf{Ensemble}}\\
          \hline
          $\sigma$ & $\epsilon$ & \textbf{Correct} & \textbf{Rejected} & \textbf{Incorrect} & $\epsilon$ & \textbf{Correct} & \textbf{Rejected} & \textbf{Incorrect} & $\epsilon$ & \textbf{Correct} & \textbf{Rejected} & \textbf{Incorrect} \\
          \hline
          0.00 & 0.05 & 57.30	& 0.00  & \textbf{42.70} & 0.10 & 37.70 & 0.00  & \textbf{62.30} & 0.30 & 16.60 & 0.00  & \textbf{83.40} \\ 
          0.30 &      & 56.80	& 0.00  & \textbf{43.20} &      & 38.50 & 0.10  & \textbf{61.40} &      & 16.90 & 0.00  & \textbf{83.10} \\ 
          0.70 &      & 42.30	& 32.70 & \textbf{25.00} &      & 26.80 & 29.50 & \textbf{43.70} &      & 7.90  & 38.20 & \textbf{53.90} \\ 
          1.00 &      & 16.80	& 77.90 & \textbf{5.30}  &      & 10.00 & 75.60 & \textbf{14.40} &      & 0.50  & 83.80 & \textbf{15.70} \\ 
          \hline
          \multicolumn{13}{|l|}{\textbf{RMAggNet}}\\
          \hline
          EC & $\epsilon$ & \textbf{Correct} & \textbf{Rejected} & \textbf{Incorrect} & $\epsilon$ & \textbf{Correct} & \textbf{Rejected} & \textbf{Incorrect} & $\epsilon$ & \textbf{Correct} & \textbf{Rejected} & \textbf{Incorrect} \\
          \hline
          0 & 0.05 & 16.20 & 75.60 & \textbf{8.20}  & 0.10 & 4.40  & 70.10 & \textbf{25.50} & 0.30 & 0.30 & 49.40 & \textbf{50.30} \\ 
          1 &      & 24.80 & 58.30 & \textbf{16.90} &      & 9.30  & 48.30 & \textbf{42.40} &      & 0.50 & 23.20 & \textbf{76.30} \\ 
          2 &      & 34.20 & 39.20 & \textbf{26.60} &      & 16.10 & 29.70 & \textbf{54.20} &      & 1.10 & 11.10 & \textbf{87.80} \\ 
          3 &      & 44.90 & 15.30 & \textbf{39.80} &      & 21.70 & 12.90 & \textbf{65.40} &      & 2.10 & 5.20  & \textbf{92.70} \\ 
          \hline
    \end{tabular}%
    }
    
    \end{subtable}
    
\end{table}
}{}
\begin{table}[!htbp]
    \caption{Results of PGD $L_\infty$ adversaries generated using open-box attacks on CIFAR-10 images with percentages of correct, rejected and incorrect classifications. Lower incorrectness is better.}
    \label{tab:cifar_direct_pgdlinf}
    \centering
    \resizebox{1\columnwidth}{!}{
    \begin{tabular}{|*{13}{c|}}
        \hline
        \multicolumn{13}{|c|}{\textbf{PGD($L_\infty$)}}\\
        \hline
        \multicolumn{13}{|l|}{\textbf{CCAT}}\\
        \hline
          $\tau$ & $\epsilon$ & \textbf{Correct} & \textbf{Rejected} & \textbf{Incorrect} & $\epsilon$ & \textbf{Correct} & \textbf{Rejected} & \textbf{Incorrect} & $\epsilon$ & \textbf{Correct} & \textbf{Rejected} & \textbf{Incorrect} \\
         \hline
          0.00 & 0.05 & 9.00 & 0.00   & \textbf{91.00} & 0.10 & 7.10 & 0.00   & \textbf{92.90} & 0.30 & 5.40 & 0.00  & \textbf{94.60} \\ 
          0.30 &      & 0.00 & 99.00  & \textbf{1.00 } &     & 0.00 & 97.50  & \textbf{2.50 } &     & 0.20 & 83.00 & \textbf{16.80} \\
          0.70 &      & 0.00 & 99.50  & \textbf{0.50 } &     & 0.00 & 98.50  & \textbf{1.50 } &     & 0.00 & 86.40 & \textbf{13.60} \\
          0.90 &      & 0.00 & 99.70  & \textbf{0.30}  &     & 0.00 & 98.80  & \textbf{1.20}  &     & 0.00 & 89.50 & \textbf{10.50} \\
          \hline
          \multicolumn{13}{|l|}{\textbf{Ensemble}}\\
          \hline
          $\sigma$ & $\epsilon$ & \textbf{Correct} & \textbf{Rejected} & \textbf{Incorrect} & $\epsilon$ & \textbf{Correct} & \textbf{Rejected} & \textbf{Incorrect} & $\epsilon$ & \textbf{Correct} & \textbf{Rejected} & \textbf{Incorrect} \\
          \hline
          0.00 & 0.05 & 1.10 & 0.00  & \textbf{98.90 }& 0.10 & 0.00 & 0.00 & \textbf{100.00} & 0.30 & 0.00 & 0.00 & \textbf{100.00} \\
          0.30 &      & 0.90 & 0.00  & \textbf{99.10 }&     & 0.00 & 0.00 & \textbf{100.00} &     & 0.00 & 0.00 & \textbf{100.00} \\
          0.70 &      & 0.50 & 2.70  & \textbf{96.80 }&     & 0.00 & 0.10 & \textbf{99.90 } &     & 0.00 & 0.00 & \textbf{100.00} \\
          1.00 &      & 0.20 & 14.80 & \textbf{85.00 }&     & 0.00 & 1.60 & \textbf{98.40 } &     & 0.00 & 0.10 & \textbf{99.90 } \\
          \hline
          \multicolumn{13}{|l|}{\textbf{RMAggNet}}\\
          \hline
          EC & $\epsilon$ & \textbf{Correct} & \textbf{Rejected} & \textbf{Incorrect} & $\epsilon$ & \textbf{Correct} & \textbf{Rejected} & \textbf{Incorrect} & $\epsilon$ & \textbf{Correct} & \textbf{Rejected} & \textbf{Incorrect} \\
          \hline
          0 & 0.05 & 9.50  & 65.20 & \textbf{25.30} & 0.10 & 3.10 & 67.60 & \textbf{29.30} & 0.30 & 0.00 & 82.30 & \textbf{17.70} \\
          1 &      & 12.80 & 43.70 & \textbf{43.50} &      & 4.00 & 44.90 & \textbf{51.10} &      & 0.00 & 50.80 & \textbf{49.20} \\
          2 &      & 15.60 & 27.20 & \textbf{57.20} &      & 5.00 & 24.10 & \textbf{70.90} &      & 0.20 & 26.20 & \textbf{73.60} \\
          3 &      & 18.50 & 12.90 & \textbf{68.60} &      & 6.40 & 9.50  & \textbf{84.10} &      & 0.20 & 11.10 & \textbf{88.70} \\
          \hline
    \end{tabular}
    }
\end{table}

\longpaper{\begin{figure}
    \begin{subfigure}{0.3\textwidth}
    \centering
    \tikzset{every picture/.style={line width=0.75pt}}
        \begin{tikzpicture}[x=0.75pt,y=0.75pt,yscale=-1,xscale=1]
        
        \draw (-45,0) node [anchor=north west][inner sep=0.75pt]    {Surrogate Model};
        \draw (0,30) node  {\includegraphics[width=1.75\linewidth]{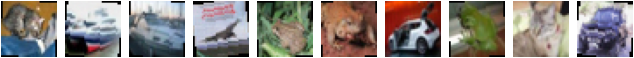}};
        \draw (0,60) node  {\includegraphics[width=1.75\linewidth]{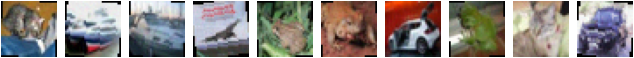}};
        \draw (0,90) node  {\includegraphics[width=1.75\linewidth]{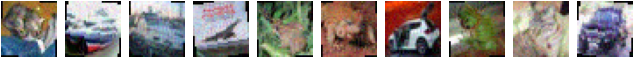}};

        \draw (255,0) node [anchor=north west][inner sep=0.75pt]    {CCAT Model};
        \draw (300,30) node  {\includegraphics[width=1.75\linewidth]{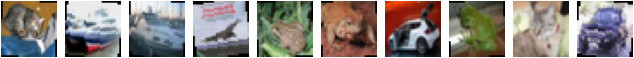}};
        \draw (300,60) node  {\includegraphics[width=1.75\linewidth]{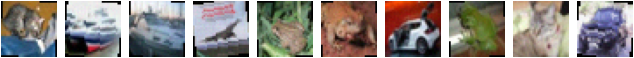}};
        \draw (300,90) node  {\includegraphics[width=1.75\linewidth]{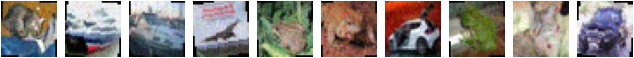}};
        
        \draw (-45,120) node [anchor=north west][inner sep=0.75pt]    {Ensemble Model};
        \draw (0,150) node  {\includegraphics[width=1.75\linewidth]{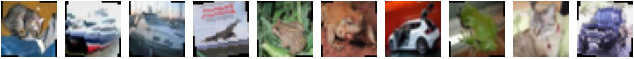}};
        \draw (0,180) node  {\includegraphics[width=1.75\linewidth]{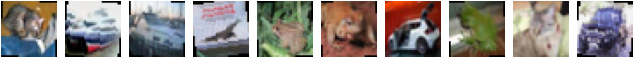}};
        \draw (0,210) node  {\includegraphics[width=1.75\linewidth]{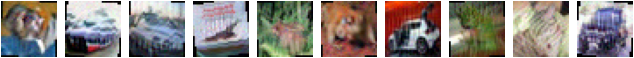}};

        \draw (255,120) node [anchor=north west][inner sep=0.75pt]    {RMAggNet Model};
        \draw (300,150) node  {\includegraphics[width=1.75\linewidth]{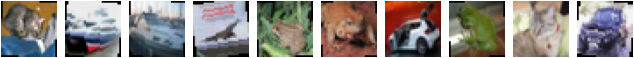}};
        \draw (300,180) node  {\includegraphics[width=1.75\linewidth]{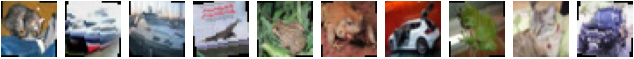}};
        \draw (300,210) node  {\includegraphics[width=1.75\linewidth]{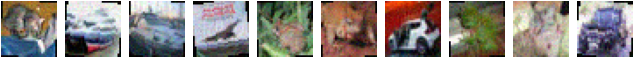}};

        \end{tikzpicture}
    
    \end{subfigure}
    \caption{Adversarial images generated using the PGD $L_2$ attack on the CIFAR-10 dataset across all models. Each architecture shows three rows of images, where each row shows the adversarial images generated at a different $\epsilon \in \{0.30, 0.75, 2.5\}$.}
    \label{fig:CIFAR_PGDL2_images}
    
\end{figure}

}{}
\longpaper{\begin{table}[htbp]
    \caption{Results for the transfer attacks using PGD $L_2$. Table \ref{tab:cifar_surrogate_pgdl2_res} shows the accuracy of the surrogate model on the PGD $L_2$ adversarial datasets. Table \ref{tab:cifar_transfer_pgdl2} shows the results of the adversarial datasets on the CCAT, Ensemble and RMAggNet models.}
    \label{tab:cifar_surrogate_transfer_res}
    \centering
    \begin{subtable}[h]{1\textwidth}
        \caption{Accuracy of the surrogate CIFAR-10 classifier on the adversarial datasets generated using PGD $L_2$ with different perturbation budgets ($\epsilon$).}
        \label{tab:cifar_surrogate_pgdl2_res}
        \begin{center}
            \begin{tabular}{|c|c|}
                \hline
                 $\epsilon$ & \textbf{Accuracy (\%)} \\
                 \hline
                 0.00 & 81.21 \\
                 0.30 & 40.30 \\
                 0.75 & 4.30  \\
                 2.50 & 0.00  \\
                 \hline
            \end{tabular}%
        \end{center}
        
    \end{subtable}
    
    \begin{subtable}[h]{1\textwidth}
    \caption{Percentage of correct, rejected and incorrect classifications of the models using transfer attacks on a surrogate CIFAR-10 classifier using the PGD $L_2$ attack. Lower incorrectness is better.}
    \label{tab:cifar_transfer_pgdl2}
    \resizebox{1\columnwidth}{!}{%
    \begin{tabular}{|*{13}{c|}}
        \hline
        \multicolumn{13}{|c|}{\textbf{PGD($L_2$)}}\\
        \hline
        \multicolumn{13}{|l|}{\textbf{CCAT}}\\
        \hline
          $\tau$ & $\epsilon$ & \textbf{Correct} & \textbf{Rejected} & \textbf{Incorrect} & $\epsilon$ & \textbf{Correct} & \textbf{Rejected} & \textbf{Incorrect} & $\epsilon$ & \textbf{Correct} & \textbf{Rejected} & \textbf{Incorrect} \\
         \hline
          0.00 & 0.30 & 67.80 & 0.00   & \textbf{32.20} & 0.75 & 48.10 & 0.00   & \textbf{51.90} & 2.5 & 8.90 & 0.00   & \textbf{91.10} \\
          0.30 &      & 49.70 & 38.60  & \textbf{11.70} &      & 4.50  & 95.10  & \textbf{0.40}  &     & 0.00 & 100.00 & \textbf{0.00}  \\
          0.70 &      & 25.70 & 71.70  & \textbf{2.60}  &      & 0.50  & 99.50  & \textbf{0.00}  &     & 0.00 & 100.00 & \textbf{0.00}  \\
          1.00 &      & 0.00  & 100.00 & \textbf{0.00}  &      & 0.00  & 100.00 & \textbf{0.00}  &     & 0.00 & 100.00 & \textbf{0.00}  \\
          \hline
          \multicolumn{13}{|l|}{\textbf{Ensemble}}\\
          \hline
           $\sigma$ & $\epsilon$ & \textbf{Correct} & \textbf{Rejected} & \textbf{Incorrect} & $\epsilon$ & \textbf{Correct} & \textbf{Rejected} & \textbf{Incorrect} & $\epsilon$ & \textbf{Correct} & \textbf{Rejected} & \textbf{Incorrect} \\
          \hline
          0.00 & 0.30 & 74.60 & 0.00  & \textbf{25.40} & 0.75 & 70.30 & 0.00  & \textbf{29.70} & 2.5 &  47.50 & 0.00  & \textbf{52.50} \\
          0.30 &      & 73.40 & 0.00  & \textbf{26.60} &      & 69.30 & 0.10  & \textbf{30.60} &     &  47.10 & 0.00  & \textbf{52.90} \\
          0.70 &      & 61.70 & 26.00 & \textbf{12.30} &      & 55.50 & 29.70 & \textbf{14.80} &     &  34.20 & 33.40 & \textbf{32.40} \\
          1.00 &      & 30.50 & 67.80 & \textbf{1.70}  &      & 25.90 & 71.20 & \textbf{2.90}  &     &  14.00 & 76.60 & \textbf{9.40}  \\
          \hline
          \multicolumn{13}{|l|}{\textbf{RMAggNet}}\\
          \hline
          EC & $\epsilon$ & \textbf{Correct} & \textbf{Rejected} & \textbf{Incorrect} & $\epsilon$ & \textbf{Correct} & \textbf{Rejected} & \textbf{Incorrect} & $\epsilon$ & \textbf{Correct} & \textbf{Rejected} & \textbf{Incorrect} \\
          \hline
          0 & 0.30 & 34.70 & 64.50 & \textbf{0.80}  & 0.75 & 24.70 & 72.80 & \textbf{2.50}  & 2.5 & 8.60  & 73.80 & \textbf{17.60} \\
          1 &      & 51.40 & 45.40 & \textbf{3.20}  &      & 40.60 & 53.60 & \textbf{5.80}  &     & 16.20 & 54.40 & \textbf{29.40} \\
          2 &      & 63.30 & 29.80 & \textbf{6.90}  &      & 50.80 & 38.90 & \textbf{10.30} &     & 23.00 & 34.40 & \textbf{42.60} \\
          3 &      & 72.40 & 15.70 & \textbf{11.90} &      & 63.30 & 16.80 & \textbf{19.90} &     & 31.40 & 14.70 & \textbf{53.90} \\
          \hline
    \end{tabular}%
    }
    
    \end{subtable}
\end{table}
}{}

\begin{table}[!htbp]
    \caption{Results of PGD $L_2$ adversaries generated using open-box attacks on CIFAR-10 images with percentages of correct, rejected and incorrect classifications. Lower incorrectness is better.}
    \label{tab:cifar_direct_pgdl2}
    \centering
    \resizebox{1\columnwidth}{!}{
    \begin{tabular}{|*{13}{c|}}
        \hline
        \multicolumn{13}{|c|}{\textbf{PGD($L_2$)}}\\
        \hline
        \multicolumn{13}{|l|}{\textbf{CCAT}}\\
        \hline
          $\tau$ & $\epsilon$ & \textbf{Correct} & \textbf{Rejected} & \textbf{Incorrect} & $\epsilon$ & \textbf{Correct} & \textbf{Rejected} & \textbf{Incorrect} & $\epsilon$ & \textbf{Correct} & \textbf{Rejected} & \textbf{Incorrect}\\
         \hline
          0.00 & 0.30 & 29.20 & 0.00   & \textbf{70.80} & 0.75 & 12.50 & 0.00   & \textbf{87.50} & 2.5 & 10.20 & 0.00   & \textbf{89.80} \\          
          0.30 &      & 0.50  & 92.10  & \textbf{7.40 } &      & 0.00  & 97.50  & \textbf{2.50 } &     & 0.00  & 99.50  & \textbf{0.50 } \\
          0.70 &      & 0.00  & 96.00  & \textbf{4.00 } &      & 0.00  & 99.20  & \textbf{0.80 } &     & 0.00  & 99.80  & \textbf{0.20 } \\
          0.90 &      & 0.00  & 97.40  & \textbf{2.60}  &      & 0.00  & 99.70  & \textbf{0.30}  &     & 0.00  & 99.80  & \textbf{0.20}  \\
          \hline
          \multicolumn{13}{|l|}{\textbf{Ensemble}}\\
          \hline
           $\sigma$ & $\epsilon$ & \textbf{Correct} & \textbf{Rejected} & \textbf{Incorrect} & $\epsilon$ & \textbf{Correct} & \textbf{Rejected} & \textbf{Incorrect} & $\epsilon$ & \textbf{Correct} & \textbf{Rejected} & \textbf{Incorrect}\\
          \hline
          0.00 & 0.30 & 54.30 & 0.00 & \textbf{45.70}  & 0.75 & 26.70 & 0.00 & \textbf{73.30}  & 2.5 & 13.40 & 0.00 & \textbf{86.60} \\
          0.30 &      & 53.30 & 0.00 & \textbf{46.70}  &      & 26.50 & 0.00 & \textbf{73.50}  &     & 13.30 & 0.00 & \textbf{86.70} \\
          0.70 &      & 42.00 & 28.60 & \textbf{29.40} &      & 23.30 & 12.90 & \textbf{63.80} &     & 13.00 & 0.90 & \textbf{86.10} \\
          1.00 &      & 23.10 & 70.20 & \textbf{6.70}  &      & 16.20 & 54.70 & \textbf{29.10} &     & 12.20 & 6.30 & \textbf{81.50} \\
          \hline
          \multicolumn{13}{|l|}{\textbf{RMAggNet}}\\
          \hline
          EC & $\epsilon$ & \textbf{Correct} & \textbf{Rejected} & \textbf{Incorrect} & $\epsilon$ & \textbf{Correct} & \textbf{Rejected} & \textbf{Incorrect} & $\epsilon$ & \textbf{Correct} & \textbf{Rejected} & \textbf{Incorrect}\\
          \hline
          0 & 0.30 & 28.70 & 69.10 & \textbf{2.20 } & 0.75 & 19.40 & 71.70 & \textbf{8.90 } & 2.5 & 5.60  & 63.00 & \textbf{31.40} \\
          1 &      & 41.20 & 52.10 & \textbf{6.70 } &      & 25.60 & 54.90 & \textbf{19.50} &     & 8.20  & 41.80 & \textbf{50.00} \\
          2 &      & 52.40 & 36.30 & \textbf{11.30} &      & 32.90 & 37.90 & \textbf{29.20} &     & 9.60  & 23.20 & \textbf{67.20} \\
          3 &      & 64.00 & 15.30 & \textbf{20.70} &      & 39.80 & 17.00 & \textbf{43.20} &     & 11.10 & 10.80 & \textbf{78.10} \\
          \hline
    \end{tabular}%
    }
\end{table}

\section{Discussion} \label{sec:discussion}

The results from Section~\ref{sec:results} allow us to determine how RMAggNet can be used, and when it may have advantages over competing methods. 

We start by discussing the hyperparameters of RMAggNet \longpaper{from Section~\ref{sec:results_hyperparams}}{(see extended paper \cite{longpaper}, Section 5.1)}. The selection of hyperparameters is dependent on the problem, with the most important aspect being the number of classes the dataset has. If a dataset has $|C|$ classes, then we require at least $\lceil \log_2 |C| \rceil$ networks to generate a unique encoding for each class. This works well for datasets such as MNIST or CIFAR-10, with ten classes each, which requires at least four networks, however, this approach is less optimal for datasets with a small number of classes. For datasets with few classes we are only able to produce $\binom{|C|}{s}$ unique networks (where $s$ is the number of classes we allow in the partitioned set) which limits the number of networks we can use and increases the probability that a random noise input will be assigned a valid class, which decreases adversarial defence.

Due to the nature of Reed-Muller codes, the number of networks must be a power of 2, therefore, the optimal number is often the power of 2 closest and exceeding the number of classes. Adding networks which exceed this power of 2 for error correction is  a costly task with quickly decreasing benefit \longpaper{(see table \ref{tab:RMA_m5r1})}{(see extended paper table 1)}, this should be done with caution.

The comparisons between RMAggNet, Ensemble and CCAT over the \longpaper{MNIST, }{}EMNIST and CIFAR-10 datasets on clean and adversarial inputs allow us to place RMAggNet in context with the other methods. The results for these tests are in sections \longpaper{\ref{sec:res_mnist},}{}  \ref{sec:res_emnist}, and \ref{sec:res_cifar} \longpaper{}{(more results available in the extended paper \cite{longpaper})}.

Results on the clean testing data (tables \longpaper{ \ref{tab:mnist_res_clean}, }{}\ref{tab:emnist_res} and \ref{tab:cifar10_res}) show that RMAggNet is able to train models which are competitive with the other architectures, equalling Ensemble for some datasets. This result shows that the RMAggNet method has minimal impact on clean dataset performance. 

\longpaper{Tables \ref{tab:noise_mnist} and \ref{tab:ood_mnist} show that RMAggNet is able to reject out-of-distribution data well, and is able to refuse to classify data which traditional neural networks would be forced to classify. We also see that it performs better on the uniform noise dataset than FMNIST. This is an expected result since FMNIST represents a semantically coherent dataset which is likely to share low-level features with the MNIST dataset the model was trained on. The error correction ability of RMAggNet does impact the percentage of correctly rejected inputs, with more error correction leading to a higher number of incorrect classifications. This presents the challenge of rejection with error correction. If we consider RMAggNet, the result on the clean dataset (table \ref{tab:mnist_res_clean}) shows that the best version uses $EC=3$, whereas in an out-of-distribution setting (e.g. table \ref{tab:noise_mnist}) a high $EC$ value leads to high incorrectness. A balance must be struck between the desired correctness, incorrectness, and rejection which is often dependent on the situation a model will be deployed in.}{}

Across the adversarial tests CCAT is able to reject the most adversarial inputs leading to nearly 0 incorrect classifications being made. However, CCAT does not attempt correction of any inputs which leads to nearly 0 correct classifications at most confidence thresholds. This even occurs when both Ensemble and RMAggNet recover over \longpaper{90\% of the labels from an adversarial attack (see table \ref{tab:mnist_direct_pgdlinf})}{80\% of the labels from an adversarial attack (table \ref{tab:emnist_direct_pgdl2})}. This approach to rejection means that CCAT is ideal for situations where any uncertainty in the correctness of a result cannot be tolerated. However, if we can allow some incorrectness, and want the option to reject, then Ensemble and RMAggNet allow us to classify many inputs correctly, which greatly reduces the reliance on downstream rejection handling, at the risk of a small amount of incorrectness. If we compare the Ensemble method with RMAggNet, over many of the datasets RMAggNet is able to outperform Ensemble with a slightly higher (or equal) number of correct classifications, and lower incorrectness for comparable correctness as it uses the reject option more effectively. This becomes more pronounced at higher $\epsilon$. The application of RMAggNet to datasets with many more classes, such as ImageNet, would be interesting future work since we have stated that $|C| \leq 2^k$ (section \ref{sec:rm_aggnet}), and this can be achieved by either adding more networks (increasing $m$) or increasing the polynomial degree ($r$) which decreases error correction ability and increases the probability of assigning random noise a class. Striking this balance would lead to interesting results regarding the applicability of RMAggNet to larger datasets.

\section{Conclusion}

In this paper we have seen how an architecture inspired by Reed-Muller codes can be used to create an effective CWR method which (to our knowledge) is the first approach combining its rejection and correction ability. The experimental results show the advantages of RMAggNet and allow us to determine situations where it can be beneficial to a system. Comparing the results of RMAggNet to CCAT, shows that CCAT is able to reject nearly 100\% of the adversarial images over all attacks and datasets we tested. However, this comes at the cost of rejecting inputs that could otherwise be classified correctly. The sensitive approach of CCAT could be detrimental to a system where the rejected inputs still need to be processed, either using more computationally expensive processes, or reviewed by a human. RMAggNet is able to achieve low incorrectness, often with higher correctness, meaning that, provided the system can accept small amounts of incorrectness, we can reduce the reliance on any downstream rejection handling. From the results, we can see that this can be a significant improvement for small perturbations on the \longpaper{MNIST/}{}EMNIST dataset\longpaper{s}{}. Comparing the results of RMAggNet to Ensemble, the results show that RMAggNet appears to be more resilient to adversarial attacks, particularly open-box attacks. This means that we can expect RMAggNet to be a better choice in a situation where we expect adversaries to be present.

\bibliographystyle{IEEEtran}
\bibliography{ref.bib}

\begin{thebibliography}{10}
\providecommand{\url}[1]{#1}
\csname url@samestyle\endcsname
\providecommand{\newblock}{\relax}
\providecommand{\bibinfo}[2]{#2}
\providecommand{\BIBentrySTDinterwordspacing}{\spaceskip=0pt\relax}
\providecommand{\BIBentryALTinterwordstretchfactor}{4}
\providecommand{\BIBentryALTinterwordspacing}{\spaceskip=\fontdimen2\font plus
\BIBentryALTinterwordstretchfactor\fontdimen3\font minus \fontdimen4\font\relax}
\providecommand{\BIBforeignlanguage}[2]{{%
\expandafter\ifx\csname l@#1\endcsname\relax
\typeout{** WARNING: IEEEtran.bst: No hyphenation pattern has been}%
\typeout{** loaded for the language `#1'. Using the pattern for}%
\typeout{** the default language instead.}%
\else
\language=\csname l@#1\endcsname
\fi
#2}}
\providecommand{\BIBdecl}{\relax}
\BIBdecl

\bibitem{chen2023symbolic}
X.~Chen, C.~Liang, D.~Huang, E.~Real, K.~Wang, Y.~Liu, H.~Pham, X.~Dong, T.~Luong, C.-J. Hsieh \emph{et~al.}, ``Symbolic discovery of optimization algorithms,'' \emph{arXiv preprint arXiv:2302.06675}, 2023.

\bibitem{tragakis2023fully}
A.~Tragakis, C.~Kaul, R.~Murray-Smith, and D.~Husmeier, ``The fully convolutional transformer for medical image segmentation,'' in \emph{Proceedings of the IEEE/CVF Winter Conference on Applications of Computer Vision}, 2023, pp. 3660--3669.

\bibitem{malware}
F.~Pierazzi, F.~Pendlebury, J.~Cortellazzi, and L.~Cavallaro, ``Intriguing properties of adversarial ml attacks in the problem space,'' in \emph{2020 IEEE Symposium on Security and Privacy (SP)}, 2020, pp. 1332--1349.

\bibitem{szegedy2013intriguing}
C.~Szegedy, W.~Zaremba, I.~Sutskever, J.~Bruna, D.~Erhan, I.~Goodfellow, and R.~Fergus, ``Intriguing properties of neural networks,'' \emph{arXiv:1312.6199}, 2013.

\bibitem{smith2018understanding}
L.~Smith and Y.~Gal, ``{Understanding measures of uncertainty for adversarial example detection},'' in \emph{34th Conference on Uncertainty in Artificial Intelligence 2018, UAI 2018}, vol.~2, mar 2018, pp. 560--569.

\bibitem{zou2023universal}
\BIBentryALTinterwordspacing
A.~Zou, Z.~Wang, J.~Z. Kolter, and M.~Fredrikson, ``Universal and transferable adversarial attacks on aligned language models,'' 2023. [Online]. Available: \url{https://github.com/llm-attacks/llm-attacks}
\BIBentrySTDinterwordspacing

\bibitem{morris2020textattack}
J.~Morris, E.~Lifland, J.~Y. Yoo, J.~Grigsby, D.~Jin, and Y.~Qi, ``Textattack: A framework for adversarial attacks, data augmentation, and adversarial training in nlp,'' in \emph{Proceedings of the 2020 Conference on Empirical Methods in Natural Language Processing: System Demonstrations}, 2020, pp. 119--126.

\bibitem{chen2019shapeshifter}
S.-T. Chen, C.~Cornelius, J.~Martin, and D.~H. Chau, ``Shapeshifter: Robust physical adversarial attack on faster r-cnn object detector,'' in \emph{Machine Learning and Knowledge Discovery in Databases: European Conference, ECML PKDD 2018, Dublin, Ireland, September 10--14, 2018, Proceedings, Part I 18}.\hskip 1em plus 0.5em minus 0.4em\relax Springer, 2019, pp. 52--68.

\bibitem{papernot2016distillation}
N.~Papernot, P.~McDaniel, X.~Wu, S.~Jha, and A.~Swami, ``Distillation as a defense to adversarial perturbations against deep neural networks,'' in \emph{2016 IEEE symposium on security and privacy (SP)}.\hskip 1em plus 0.5em minus 0.4em\relax IEEE, 2016, pp. 582--597.

\bibitem{goodfellow2014explaining}
I.~J. Goodfellow, J.~Shlens, and C.~Szegedy, ``Explaining and harnessing adversarial examples,'' \emph{arXiv preprint arXiv:1412.6572}, 2014.

\bibitem{verma2019error}
G.~Verma and A.~Swami, ``Error correcting output codes improve probability estimation and adversarial robustness of deep neural networks,'' \emph{Advances in Neural Information Processing Systems}, vol.~32, 2019.

\bibitem{cortes2016learning}
C.~Cortes, G.~DeSalvo, and M.~Mohri, ``Learning with rejection,'' in \emph{Algorithmic Learning Theory: 27th International Conference, ALT 2016, Bari, Italy, October 19-21, 2016, Proceedings 27}.\hskip 1em plus 0.5em minus 0.4em\relax Springer, 2016, pp. 67--82.

\bibitem{charoenphakdee2021classification}
N.~Charoenphakdee, Z.~Cui, Y.~Zhang, and M.~Sugiyama, ``Classification with rejection based on cost-sensitive classification,'' in \emph{International Conference on Machine Learning}.\hskip 1em plus 0.5em minus 0.4em\relax PMLR, 2021, pp. 1507--1517.

\bibitem{song2021error}
Y.~Song, Q.~Kang, and W.~P. Tay, ``Error-correcting output codes with ensemble diversity for robust learning in neural networks,'' in \emph{Proceedings of the AAAI Conference on Artificial Intelligence}, vol.~35, no.~11, 2021, pp. 9722--9729.

\bibitem{papernot2016transferability}
N.~Papernot, P.~McDaniel, and I.~Goodfellow, ``Transferability in machine learning: from phenomena to black-box attacks using adversarial samples,'' \emph{arXiv preprint arXiv:1605.07277}, 2016.

\bibitem{Stutz2020ICML}
D.~Stutz, M.~Hein, and B.~Schiele, ``Confidence-calibrated adversarial training: Generalizing to unseen attacks,'' \emph{Proceedings of the International Conference on Machine Learning {ICML}}, 2020.

\bibitem{gamalSA}
A.~Gamal, L.~Hemachandra, I.~Shperling, and V.~Wei, ``Using simulated annealing to design good codes,'' \emph{IEEE Transactions on Information Theory}, vol.~33, no.~1, pp. 116--123, 1987.

\bibitem{mullerRM}
D.~E. Muller, ``Application of boolean algebra to switching circuit design and to error detection,'' \emph{Transactions of the I.R.E. Professional Group on Electronic Computers}, vol. EC-3, no.~3, pp. 6--12, 1954.

\bibitem{reedRM}
I.~Reed, ``A class of multiple-error-correcting codes and the decoding scheme,'' \emph{Transactions of the IRE Professional Group on Information Theory}, vol.~4, no.~4, pp. 38--49, 1954.

\bibitem{hammingCode}
R.~W. Hamming, ``Error detecting and error correcting codes,'' \emph{The Bell System Technical Journal}, vol.~29, no.~2, pp. 147--160, 1950.

\bibitem{carlini2019evaluating}
N.~Carlini, A.~Athalye, N.~Papernot, W.~Brendel, J.~Rauber, D.~Tsipras, I.~Goodfellow, A.~Madry, and A.~Kurakin, ``On evaluating adversarial robustness,'' \emph{arXiv preprint arXiv:1902.06705}, 2019.

\bibitem{madry2017towards}
A.~Madry, A.~Makelov, L.~Schmidt, D.~Tsipras, and A.~Vladu, ``Towards deep learning models resistant to adversarial attacks,'' \emph{arXiv preprint arXiv:1706.06083}, 2017.

\bibitem{brendel2018decisionbased}
W.~Brendel, J.~Rauber, and M.~Bethge, ``Decision-based adversarial attacks: Reliable attacks against black-box machine learning models,'' in \emph{International Conference on Learning Representations}, 2018.

\bibitem{cohen2017emnist}
G.~Cohen, S.~Afshar, J.~Tapson, and A.~Van~Schaik, ``Emnist: Extending mnist to handwritten letters,'' in \emph{2017 international joint conference on neural networks (IJCNN)}.\hskip 1em plus 0.5em minus 0.4em\relax IEEE, 2017, pp. 2921--2926.

\bibitem{xiao2017fmnist}
H.~Xiao, K.~Rasul, and R.~Vollgraf. (2017) Fashion-mnist: a novel image dataset for benchmarking machine learning algorithms.

\bibitem{rauber2017foolboxnative}
\BIBentryALTinterwordspacing
J.~Rauber, R.~Zimmermann, M.~Bethge, and W.~Brendel, ``Foolbox native: Fast adversarial attacks to benchmark the robustness of machine learning models in pytorch, tensorflow, and jax,'' \emph{Journal of Open Source Software}, vol.~5, no.~53, p. 2607, 2020. [Online]. Available: \url{https://doi.org/10.21105/joss.02607}
\BIBentrySTDinterwordspacing

\bibitem{rauber2017foolbox}
J.~Rauber, W.~Brendel, and M.~Bethge, ``Foolbox: A python toolbox to benchmark the robustness of machine learning models,'' in \emph{Reliable Machine Learning in the Wild Workshop, 34th International Conference on Machine Learning}, 2017.

\bibitem{athalye2018obfuscated}
A.~Athalye, N.~Carlini, and D.~Wagner, ``Obfuscated gradients give a false sense of security: Circumventing defenses to adversarial examples,'' in \emph{International conference on machine learning}.\hskip 1em plus 0.5em minus 0.4em\relax PMLR, 2018, pp. 274--283.

\bibitem{dong2018boosting}
Y.~Dong, F.~Liao, T.~Pang, H.~Su, J.~Zhu, X.~Hu, and J.~Li, ``Boosting adversarial attacks with momentum,'' in \emph{Proceedings of the IEEE conference on computer vision and pattern recognition}, 2018, pp. 9185--9193.

\bibitem{he2016deep}
K.~He, X.~Zhang, S.~Ren, and J.~Sun, ``Deep residual learning for image recognition,'' in \emph{Proceedings of the IEEE conference on computer vision and pattern recognition}, 2016, pp. 770--778.

\bibitem{jeevan2022wavemix}
P.~Jeevan, K.~Viswanathan, and A.~Sethi, ``Wavemix-lite: A resource-efficient neural network for image analysis,'' \emph{arXiv preprint arXiv:2205.14375}, 2022.

\end{thebibliography}
\end{document}